\documentclass[letterpaper,twocolumn]{article} 
\usepackage{times}  
\usepackage{helvet}  
\usepackage{courier}
\usepackage[hyphens]{url} 
\usepackage{graphicx}
\usepackage{caption} 
\usepackage{algorithm}
\usepackage{algorithmic}
\usepackage{newfloat}
\usepackage{listings}
\usepackage{geometry}
\floatstyle{ruled}
\newfloat{listing}{tb}{lst}{}
\floatname{listing}{Listing}
\usepackage[utf8]{inputenc}
\usepackage[english]{babel}
\usepackage{enumerate}
\usepackage{ amssymb}
\usepackage{amsmath}
\usepackage{amsthm}
\usepackage{ dsfont }
\usepackage{xspace}
\usepackage{subcaption}
\usepackage{etoolbox}
\title{Eliciting Kemeny Rankings}
\author{
  Anne-Marie George\textsuperscript{\rm 1},
  Christos Dimitrakakis\textsuperscript{\rm 1, \rm 2} \\
\normalsize
    \textsuperscript{\rm 1}University of Oslo, Norway\\
    \textsuperscript{\rm 2}University of Neuchatel, Switzerland\\
    annemage@uio.no, christos.dimitrakakis@unine.ch
}
\date{}

\newtheorem{example}{Example}

\newtheorem{lemma}{Lemma}
\newtheorem{corollary}{Corollary}

\newtheorem{proposition}{Proposition}

\usepackage[sort&compress]{cleveref}
\crefname{construction}{Construction}{Constructions}
\crefname{claim}{Claim}{Claims}
\crefname{paragraph}{Paragraph}{Paragraphs}
\crefname{observation}{Observation}{Observations}
\crefname{theorem}{Theorem}{Theorems}
\crefname{lemma}{Lemma}{Lemmata}
\crefname{proposition}{Proposition}{Propositions}
\crefname{corollary}{Corollary}{Corollaries}
\crefname{remark}{Remark}{Remarks}
\crefname{section}{Section}{sections}
\crefname{chapter}{Chapter}{Chapters}
\crefname{figure}{Figure}{Figures}
\crefname{table}{Table}{Tables}
\crefname{definition}{Definition}{Definitions}
\crefname{algorithm}{Algorithm}{Algorithms}
\crefname{equation}{Equation}{Equations}
\crefname{appendix}{Appendix}{Appendices}

\newcommand{\appsymb}{$\bigstar$}
\newcommand{\appref}[1]{{\appsymb}}
\def\appendixProofText{}
\newcommand{\appendixsection}[1]{%
  \gappto{\appendixProofText}{\section{Additional Material for Section~\ref{#1}}\label{app:#1}}
}
\newcommand{\toappendixx}[1]{}
\newcommand{\toappendix}[1]{%
\gappto{\appendixProofText}
  {{
    #1
  }}
}

\newcommand{\appendixproof}[2]{%
  \gappto{\appendixProofText}
  {
    \subsection{Proof of \cref{#1}}\label{proof:#1}
    #2
  }
}

\newcommand{\kt}{\text{d}_\text{Kt}} 
\newcommand{\argmin}{\text{argmin}} %
\newcommand{\kemeny}{\text{K}} 
\newcommand{\kscore}{\text{KS}} 
\newcommand{\arms}{\text{k}} 

\newcommand{\winmatrix}{\mathcal{Q}(\arms)} 
\newcommand{\prefmatrix}{\mathcal{P}(\arms,n)} 

\begin{document}

 \newgeometry{
 left=0.75in,
 right=0.75in,
 top=1.25in,
 bottom=1.25in
 }
\maketitle

\begin{abstract}
We formulate the problem of eliciting agents' preferences with the goal of finding a Kemeny ranking as a Dueling Bandits problem.
Here the bandits' arms correspond to alternatives that need to be ranked and the feedback corresponds to a pairwise comparison between alternatives by a randomly sampled agent.
We consider both sampling with and without replacement, i.e., the possibility to ask the same agent about some comparison multiple times or not.

We find approximation bounds for Kemeny rankings dependant on confidence intervals over estimated winning probabilities of arms. 
Based on these we state algorithms to find Probably Approximately Correct (PAC) solutions and elaborate on their sample complexity for sampling with or without replacement.
Furthermore, if all agents' preferences are strict rankings over the alternatives, we provide means to prune confidence intervals and thereby guide a more efficient elicitation. We formulate several adaptive sampling methods that use look-aheads to estimate how much confidence intervals (and thus approximation guarantees) might be tightened. 
All described methods are compared on synthetic data.
\end{abstract}

\section{Introduction}

\footnote[0]{This is a long version of the AAAI'24 publication under the same title.}
Decision making based on the preferences of a group of individuals' is tackled in various settings by AI systems. 
To ensure a good societal outcome, the fair aggregation of the society members' preferences over alternatives plays a central role. 
Many such aggregation functions for agents' or voters' preferences 
have been proposed by politicians, mathematicians, economists and computer scientists.
Research in the area of Social Choice and more specifically Voting Theory analyses and develops such aggregation functions algorithmically and axiomatically~\cite{BCE+16}. 
While impossibility results force all such functions to be flawed by computational complexity or violation of fairness guarantees, e.g.~\cite{Arr50,Sat75}, many are well researched and applied in practise. 
However, it is usually assumed that voters specify their  preferences completely before the desired aggregation can be determined. This can be excessively time consuming or costly for voters, especially for large numbers of alternatives as in movie or online-shopping platforms. 
Next we describe our contributions for efficiently eliciting aggregated rankings 
under Kemeny's rule, 
related work and outline  of this paper.

\subsection{Contributions and Related Work}

We formulate the problem of eliciting voter preferences as a Dueling Bandit problem~\cite{BBEH21} in which arms of bandits correspond to alternatives, and the bandit feedback for pulling two arms corresponds to a random voter's preference over the two arms. This feedback is then summarised in a matrix of winning probabilities over arms.
Our goal is to elicit an aggregated ranking of voter preferences based on some voting rule. Here, the duelling bandit formulation is applicable for any voting rule that is a \emph{C2 function} according to Fishburn's classification of preference aggregation functions~\cite{Fish77}. 
That is, functions that only require the margin of voters that prefer one alternative over another as input, and not the exact combinations of pairwise preferences the voters have. 
For such functions, we can thus assume the input to be a \textit{preference matrix} indicating the fraction of voters preferring an alternative over another for all pairs of alternatives.
The results by~\cite{BFNS22} imply that it is NP-hard to decide whether a given matrix 
corresponds 
to a set of voters' preference orders (i.e., linear orders) over the alternatives. It follows that it is not possible to find a characterisation of such matrices that can be checked in polynomial time. Nevertheless, we identify properties of such matrices.\\
\textbf{Contribution 1:} In Section~\ref{sec: preliminaries}, we connect the vote elicitation and dueling bandit frameworks for C2 functions, and propose three properties of preference matrices, which will help devising adaptive elicitation strategies:  Completeness, Triangle Inequality and Realisable Borda Scores.\\

Next, we focus our attention on the \textit{voting rule} devised by John Kemeny in 1959~\cite{Kem59}, which is a preference aggregation function with compelling properties such as monotonicity and Condorcet winner consistency~\cite{YoLe78}.
The output of this rule is a set of rankings over the alternatives which minimise the \textit{Kemeny score}.
The Kemeny score measures voter disagreement in terms of pairwise disagreement over the ranked alternatives, i.e., Kendall's tau distance~ \cite{Ken38}.
In general, it is NP-hard to compute Kemeny rankings~\cite{BTT89} but polynomial time solvable for specific types of preferences, such as single-peaked ones~\cite{BBHH15}.
A perk of Kemeny's rule, as a C2 function, is that it can be readily applied to any matrix. That is, even when voters only specify parts of their preferences, i.e., an incomplete order over the alternatives, or when their preferences are non-transitive. The corresponding matrix in that case only contains fractions of voters that specified their preference over two alternatives.
This enables us to compute an approximate Kemeny ranking even when not all preferences have been elicited yet. 
Our goal then is to sample pairs of arms (i.e., query voters to compare two alternatives) such that the approximate Kemeny ranking based on the sample means is close (for some distance measure) to an optimal Kemeny ranking. 
Note that while approximation algorithms exist for Kemeny rankings, e.g., ~\cite{BFG+08}, the approximation bounds usually concern the approximation w.r.t. a given instance. Contrarily, we consider the approximation of Kemeny rankings w.r.t. a similar instance while applying an exact method.\\
\textbf{Contribution 2:} In Section~\ref{sec: kemenys rule}, we formulate approximation bounds for (1) Kendall tau distance and (2) Kemeny Score difference, between the true Kemeny ranking of a matrix of winning probabilities and the Kemeny ranking of an 
approximate matrix.
These bounds are based on confidence intervals around sample means for the winning probabilities.\\

Note that bandit problems with the goal of finding a ranking over the arms (as opposed to one or several winning arms) have been analysed for the C2 functions Borda and Copeland, see e.g.~\cite{BSBH14,FOPS17,LiLu18}, and other winner concepts such as von Neumann winners~\cite{BKSZ16}.
These works establish \textit{probably approximately correct} (PAC) algorithms and analyse their sample complexity. To the best of our knowledge such considerations do not exist for Kemeny rankings. In fact,~\cite{BBEH21} propose Kemeny rankings for future work. 
Also, no related work considers sampling from a fixed population without replacement (as would be reasonable when given a set of voters which shall not be queried twice about the same pair of alternatives).
What is more, approximations of rankings w.r.t. Kemeny scores (i.e., agents' disagreement) have not been considered under any aggregation function so far.\\
\textbf{Contribution 3:} In Section~\ref{sec: PAC algos and sampling complexity}, we formulate PAC algorithms for Kemeny rankings w.r.t. Kemeny scores of a matrix of winning probabilities for two distinct cases of sampling methods: (1) sampling bandit feedback from voters u.a.r., i.e., sampling with replacement, (2) sampling bandit feedback from a hypergeometric distribution, i.e., sampling without replacement, meaning that voters cannot be asked multiple times about the same pair of alternatives. 
These PAC algorithms are based on our approximation results (see Contribution 2) and do not assume any properties for the matrix of winning probabilities.
We give concrete sample complexities for desired approximation values and probabilities. \\
\textbf{Contribution 4:} 
In Section~\ref{sec: pruning confidence intervals} we show how to prune confidence intervals for winning probabilities of arms based on the properties of preference matrices (see Contribution 1), i.e., for the case that the matrix originates from a set of (linear) preference orders.
We discuss several adaptive sampling methods that use look-aheads to anticipate possible effects of pruning.
Finally, we analyse the sample complexity experimentally on synthetic data for all developed methods.

\section{Voting and Bandits}\label{sec: preliminaries}
\appendixsection{sec: preliminaries}
In a voting setting, we assume to have a set $N$ of voters and a set $C=[\arms]$ of $\arms$ candidates or alternatives. 
Let $\mathcal{L}(C)$ be the set of all linear orders or rankings over $C$.
We usually consider the preferences of the voters over the alternatives to be linear orders $\succ_v \in \mathcal{L}(C)$ for $v\in N$. A typical goal in this setting is to aggregate the preferences of voters into an outcome ranking or to determine a winner. A social preference function is a function that maps a \emph{preference profile} $P=\{\succ_v \mid v\in N\}$ to one or several output rankings. One example of such a rule is Kemeny's rule, which aims to minimise the disagreement of the voters with the output ranking.

In a duelling bandits setting, we have a set $C = [\arms]$ of $\arms$ arms of bandits. In every time step, we pull two arms yielding feedback of which 
one
is better.
That is, when pulling arms $i$ and $j$ we get feedback $i \succ j$ with some (unknown) probability $q_{ij}$ and 
feedback $j \succ i$ with  inverse probability $q_{ji} = 1 - q_{ij}$. The winning probabilities can be summarised in a matrix $Q \in [0,1]^{\arms \times \arms}$ where by convention $q_{ii} = 0.5$ for all $i \in [\arms]$. 
Let $\mathcal{Q}(\arms) = \{Q \in [0,1]^{\arms \times \arms} \mid q_{ji} = 1 - q_{ij} \; \text{and} \; {q_{ii} = 0.5} \; \forall  i,j\in [\arms]\}$ be the set of such matrices. 
We assume  the true matrix of winning probabilities is unknown and must be approximated, with the ultimate goal of finding a good ranking of the arms.
This goal can be expressed by a measure of regret and literature in this field usually analyses regret bounds in the number of samples.

To connect the two settings, we make the following observation: A preference profile $P \in \mathcal{L}(C)^{|N|}$ of voters $N$ over alternatives $C=[\arms]$  can be translated to a preference matrix $Q \in \winmatrix$ by defining $q_{ij} = \frac{|\{v \in N \mid i \succ_v j\}|}{|N|}$ and $q_{ii} = 0.5$ for all $i,j \in [\arms]$. Thus, the  alternatives are seen as  arms of bandits and the winning probabilities $q_{ij}$ are equal to the fraction of voters preferring alternative $i$ over $j$. We denote the set of such \textit{preference matrices} that are originating from preference profiles by $\prefmatrix$, where $\arms$ is the number of alternatives and $n$ the number of voters. Because voters' preferences  are linear orders, and transitive, $\prefmatrix \varsubsetneq \winmatrix$.
We show further properties of such matrices. 
\begin{lemma}[\appref{le: cond-for-pref-matrix}$^2$]\label{le: cond-for-pref-matrix}
Let $P=\{\succ_v \in\mathcal{L}([\arms]) \mid v \in [n]\}$ be a profile of $n$ voters and $\arms$ alternatives. Then $Q \in \prefmatrix$ with $q_{ij} = \frac{1}{n}|\{v \in [n] \mid i \succ_v j\}|$ and $q_{ii} = 0.5$ satisfies:  \begin{enumerate}
    \item $q_{ij} \cdot n \in [n]$ and $q_{ij} + q_{ji} = 1$ for all distinct $i,j \in [\arms]$.\\ 
    \phantom{dummy}\hfill (Completeness)
    \item $q_{lj} + q_{ji}\geq q_{li}$ for all distinct $i,j,l \in [\arms]$.\\ 
    \phantom{dummy}\hfill (Triangle Inequality)
    \item $\sum_{i \in A} \sum_{j \in [\arms]\setminus \{i\}} q_{ij} \leq  \frac{1}{n} \cdot |A|\cdot (\arms - \frac{|A|+1}{2})$\\ for all $A \subseteq [\arms]$.  \hfill (Realisable Borda Scores)
\end{enumerate}
\end{lemma}
\footnote{Proofs of results marked with \appref{} are deferred to the appendix.}
\appendixproof{le: cond-for-pref-matrix}{
\begin{proof}
The first property follows directly from the definition of $Q$ and the fact that voter preferences are strict orders and illustrates that $\prefmatrix \subseteq \winmatrix$.

Let us consider the second property.
To prove this, denote by $q_{i>j>l}$ the fraction of voters $v$ that prefer $i \succ_v j \succ_v l$, and similar for other orders. There are only six possible ways for the voters to order three distinct arms $i,j,l$ and always three of these orders reflect all possibilities of ranking two arms in a specific order. We have, e.g., $q_{lj} = q_{i>l>j} + q_{l > i > j} + q_{l > j > i}$. Then 
\begin{align*}
    &q_{lj} \qquad\geq \quad q_{li} - q_{ji}\\
    \Leftrightarrow \quad 
    &q_{i>l>j} + q_{l > i > j}+ q_{l > j > i} \\
    &\qquad\quad\geq \quad q_{j > l > i} + q_{l > j > i} + q_{l > i > j} \\
    &\qquad\qquad\; - q_{l > j > i} - q_{j > l > i} - q_{j > i > l}\\
   \Leftrightarrow \quad &q_{i>l>j} + q_{l > j > i}\\
   &\qquad\quad\geq \;- q_{j > i > l},
\end{align*}
which is trivially true.

For the third property, consider the Borda score of an arm / alternative $i \in [\arms]$ which, for a given preference profile $P$, is defined as $B(i) := \sum_{v\in [n]} |\{j \in [\arms]\setminus \{i\} \mid i \succ_v j\}| = \sum_{j\in [\arms]\setminus \{i\}} |\{v \in [n] \mid i \succ_v j\}| = n\cdot \sum_{j\in [\arms]\setminus \{i\}} q_{ij}$.
That is, the Borda score for an alternative $i$ counts the position of every occurrence of $i$ in the preference rankings of the voters. For example, if $i$ is top ranked in voter $v$'s preference ranking, this contributes $\arms-1$ to the Borda score of $i$ because there are $\arms-1$ other alternatives that are ranked lower than $i$.
Any set of alternatives $A \subseteq [\arms]$ can at most be ranked at the top $|A|$ positions of the voters rankings in a preference profile and thus achieving at most a combined Borda score of $\sum_{i \in A} B(i) \leq \sum_{x=1, \dots, |A|} \arms-x = |A|\cdot \arms - \frac{|A|\cdot (|A|+1)}{2} = |A| \cdot (\arms - \frac{|A|+1}{2})$. Thus, $\sum_{i \in A} \sum_{j\in [\arms]\setminus \{i\}} q_{ij} = \frac{1}{n} \sum_{i \in A} B(i) \leq  \frac{1}{n} \cdot |A| \cdot (\arms - \frac{|A|+1}{2})$
for all $A \subseteq [\arms]$.
\end{proof}
}

The next example shows that mapping preference profiles to preference matrices is not injective. 
\begin{example}
    Consider the two preference profiles $P_1 ,P_2$ over three alternatives below. 
\[P_1 = \{1 \succ 2 \succ 3, 1 \succ 2 \succ 3, 3 \succ 2 \succ 1\}\]
\[P_2 = \{1 \succ 2 \succ 3, 2 \succ 1 \succ 3, 3 \succ 1 \succ 2\}.\]
The two profiles  map to the same preference matrix $Q$:
\[Q = \begin{bmatrix}
\frac{1}{2} & \frac{2}{3} & \frac{2}{3} \\
\frac{1}{3} & \frac{1}{2} & \frac{2}{3} \\
\frac{1}{3} & \frac{1}{3} & \frac{1}{2} 
\end{bmatrix} . \]
\end{example}

\section{Kemeny's Rule}\label{sec: kemenys rule}
\appendixsection{sec: kemenys rule}

Kemeny's Rule outputs (after applying some tie-breaking) a strict ranking of alternatives / arms that minimises the disagreement of the voters. To measure disagreement of voters with a ranking we can apply Kendall's tau distance (or ``inversion distance''). The Kendall-tau distance between two preference orders $\succ_v, \succ_w \in \mathcal{L}(C)$ is defined as follows (where $\mathds{I}$ is the indicator function):
\begin{align*}
    \kt(\succ_v, \succ_w) &= \sum\nolimits_{\{i,j\} \subseteq C} \mathds{I}(i \succ_v j)\cdot \mathds{I}(j \succ_w i) \\
    & = \sum\nolimits_{(i,j) \in \succ_v} \mathds{I}(j \succ_w i).
\end{align*}
The Kemeny score of a ranking $\succ \in \mathcal{L}(C)$ given the preference rankings of voters $N$ is defined as 
$\kscore(\{\succ_v\}_{v \in N},\succ) = \sum\nolimits_{v \in N} \kt(\succ,\succ_v)$
and the set of Kemeny rankings (before tie-breaking) is  
$\kemeny(\{\succ_v\}_{v \in N}) = \argmin_{\succ \in \mathcal{L}(C)} \kscore(\{\succ_v\}_{v \in N},\succ)$.
In the same spirit of measuring disagreement, we can define the Kemeny rankings for a given matrix $Q \in [0,1]^{\arms \times \arms}$, where $q_{ij}$ is the probability that arm $i$ beats arm $j$. The Kemeny score of a ranking $\succ \in \mathcal{L}(C)$ with $C = [\arms]$ is proportional to the probability that the outcome of any random sample of a pair of arms is not aligned with $\succ$, i.e., $\kscore(Q,\succ) = \sum\nolimits_{(i,j) \in \succ} q_{ji}$
and the set of Kemeny rankings is  $\kemeny(Q) = \argmin_{\succ \in \mathcal{L}(C)} \kscore(Q,\succ)$.

In general, Kemeny's rule is not resolute, i.e., the minimum Kemeny score may be assumed by several rankings. A tie-breaking rule allows to choose one among the rankings that minimise the Kendall-tau distance. In this paper, we assume to break ties deterministically 
whenever necessary.

Given only an approximation of a preference matrix $Q$, we can also only approximate its Kemeny ranking. To measure the quality of such an approximation, two natural distance measures for such rankings can be applied: the Kendall-tau distance between the true and the approximate ranking, or the difference in their Kemeny scores w.r.t. the true matrix $Q$. In the following, we analyse approximation guarantees based on both measures. While Kemeny's rule is NP hard to compute, there exist polynomial time approximation algorithms~\cite{BFG+08}. For an approximate Kemeny ranking of an approximate matrix, we can simply combine the approximation bounds.
Assume matrix $Q$ has Kemeny ranking $\tau$ and its approximation matrix $\hat{Q}$ has Kemeny ranking $\hat{\tau}$ that is a $\rho$-approximation w.r.t. Kemeny-scores, i.e.,  $\kscore(Q,\hat{\tau}) \leq \kscore(Q,\tau) + \rho$.  Then an additive $\alpha$-approximation $\dot{\tau}$ of $\hat{\tau}$, i.e., 
$\kscore(\hat{Q},\dot{\tau}) \leq \kscore(\hat{Q},\hat{\tau}) + \alpha$, yields $\kscore(Q,\dot{\tau}) < \kscore(Q,\tau) + \rho + \alpha$. Similarly, a multiplicative $\lambda$-approximation $\dot{\tau}$ with 
$\kscore(\hat{Q},\dot{\tau}) \leq (1+\lambda)\kscore(\hat{Q},\hat{\tau})$ yields $\kscore(Q,\dot{\tau}) < (1+\lambda)\kscore(Q,\tau) + (1+\lambda)\rho$.

\subsection{Kendall-tau Distance of Approx. Kemeny Rankings}
Consider two rankings $\tau, \hat{\tau} \in \mathcal{L}(C)$ with $|C|=\arms$. If the two rankings are equal, their Kendall-tau distance is 0; if they are reversed, their Kendall-tau distance is maximal, i.e., $\frac{(\arms - 1)\arms}{2}$. More generally, the Kendall-tau distance of any two rankings $\tau, \hat{\tau} \in \mathcal{L}(C)$ with $|C|=\arms$, is an integer in $\{0, \dots, \frac{(\arms - 1)\arms}{2}\}$. 
The following lemma shows that the Kendall-tau distance of the Kemeny rankings $\kemeny(Q)$ and $\kemeny(\hat{Q})$ of a preference matrix $Q$ and a close approximation $\hat{Q}$ of $Q$ can be maximal.
This is proven by construction of an example with values in $Q$ that are very close to $0.5$.
Thus, no matter the number of samples, 
when considering the Kemeny ranking based on the sample means, we cannot guarantee any improvement in terms of Kendall-tau distance.

\begin{lemma}[\appref{le: kt-dist-sampling-is-bad}]\label{le: kt-dist-sampling-is-bad}
For any small\footnote{For the construction of our proof, we require $\epsilon < \frac{(\arms - 1)\arms}{2}$.} $\epsilon >0$, there exist matrices $Q, \hat{Q} \in \winmatrix$ with $\|Q-\hat{Q}\|_1 =\epsilon$ and $\tau = \kemeny(Q)$ and $\hat{\tau} = \kemeny(\hat{Q})$ such that $\kt(\tau, \hat{\tau}) = \frac{(\arms - 1)\arms}{2}$.
\end{lemma}
\appendixproof{le: kt-dist-sampling-is-bad}{
\begin{proof}
Let $\epsilon >0$ be arbitrarily small and let $Q \in \winmatrix$ be the matrix given by 
\begin{align*}
    \hat{q}_{ij} = 
    \begin{cases} 
    \frac{1-\delta}{2} &\text{ if } j < i\\
    \frac{1+\delta}{2} &\text{ if } j > i
    \end{cases}
\end{align*}
where $\delta = \frac{\epsilon}{\arms(\arms-1)}$.
Furthermore, let $\hat{Q} = Q'$ be the transpose of matrix $Q$. 
Then $\|Q-\hat{Q}\|_1 =\delta \cdot \arms(\arms-1) = \epsilon$.
The unique Kemeny ranking for $Q$ is given by the ranking $\tau = 1 \succ 2 \succ \dots \succ \arms$. Because of symmetry, the only Kemeny ranking for $\hat{Q}$ is the reverse ranking $\hat{\tau} = \arms \succ \arms-1 \succ \dots \succ 1$. Thus, the Kendall-tau distance between $\tau$ and $\hat{\tau}$ is $\kt(\tau, \hat{\tau}) = \frac{(\arms - 1)\arms}{2}$.
\end{proof}
}

If the matrix of winning probabilities is indeed a preference matrix, i.e., reflecting fractions of $n$ voters, then a reasonable restriction is to ask every voter at most once about their preference over a pair of arms. Disregarding inference reached through transitivity of voter preferences, a total of $n\cdot\frac{(\arms - 1)\arms}{2}$ queries is necessary to recover the true preference matrix.
However, 
if values in the preference matrix are close to $0.5$,
even asking $(n-1)\cdot\frac{(\arms - 1)\arms}{2}$ queries can still lead to an approximate Kemeny ranking with maximal Kendall-tau distance to the true Kemeny ranking. 

\begin{lemma}[\appref{le: kt-dist-is-no-good-for-pref-matrix}]\label{le: kt-dist-is-no-good-for-pref-matrix}
For any number of arms $\arms > 2$ and number of voters $n>2$, there exist preference profiles with corresponding preference matrices $Q \in \prefmatrix$ and $\hat{Q} \in \mathcal{P}(\arms,n-1)$ with $\|Q-\hat{Q}\|_1 = \frac{(\arms - 1)\arms}{2(n-1)}$ and $\tau = \kemeny(Q)$ and $\hat{\tau} = \kemeny(\hat{Q})$ such that $\kt(\tau, \hat{\tau}) = \frac{(\arms - 1)\arms}{2}$.
\end{lemma}
\appendixproof{le: kt-dist-is-no-good-for-pref-matrix}{
\begin{proof}
First, consider an even number of voters $n$. 
Assume half of the voters have preference $\tau = 1 \succ 2 \succ \dots \succ \arms$ and the other half has preference $\hat{\tau} = \arms \succ \arms-1 \succ \dots \succ 1$.
The corresponding preference matrix $Q$ is given by $q_{ij} = 0.5$ for all $i,j \in [\arms]$, and the Kemeny ranking, due to our tie-breaking rule, is $\kemeny(Q) = \tau$. 
Now consider the preference profile corresponding to the preferences of all voters except one of the voters with preference $\tau$. The corresponding preference matrix $\hat{Q}$ is given by 
\begin{align*}
    \hat{q}_{ij} = 
    \begin{cases} 
    \frac{n}{2(n-1)} &\text{ if } j < i\\
    \frac{n-2}{2(n-1)} &\text{ if } j > i
    \end{cases}
\end{align*}
and the Kemeny ranking is $\kemeny(\hat{Q}) = \hat{\tau}$.
Thus $\|Q-\hat{Q}\|_1 = \frac{(\arms - 1)\arms}{2(n-1)}$ and $\kt(\tau, \hat{\tau}) = \frac{(\arms - 1)\arms}{2}$.

For an uneven number of voters $n$, we can construct a similar example with $n_* = \lfloor \frac{n}{2} \rfloor$ voters preferring $\tau$ and $n^* = \lceil \frac{n}{2} \rceil$ voters preferring  $\hat{\tau}$, such that the Kemeny ranking is $\hat{\tau}$. Then, by eliminating preferences of one of the $n^*$ voters, we create a preference profile with Kemeny ranking $\tau$ (due to tie-breaking).

While this proof uses tie-breaking based on ranking $\tau$, similar examples can be constructed for any other deterministic tie-breaking rule.
\end{proof}
}

This shows that especially for large numbers of voters, the approximation error for approximating matrix $Q$ could be very small, while the Kendall-tau distance is maximal.
A similar result can be obtained when querying voters about all but one pair of alternatives.
Consider a Condorcet cycle for which $q_{12} = q_{23} = q_{31} = 2/3$. Then the tie-breaking rule, e.g., $1 \succ 2 \succ 3$, coincides with the Kemeny ranking with a score of $4/3$. When eliciting all preferences over all pairs of arms but $(2,3)$, and initialising $q_{23} = q_{32} = 0.5$, the approximated Kemeny ranking is $3 \succ 2 \succ 1$, independent of tie-breaking, yielding a maximal KT-distance.

The results in this section are based on examples with matrices / preference profiles for which Kemeny rankings have very high Kemeny scores.
Thus, while potentially yielding very different rankings, the similar matrices in the proofs do not yield much difference in terms of voter satisfaction by their corresponding Kemeny rankings.
In the next section, we concentrate on bounds of Kemeny score differences between Kemeny rankings of similar matrices.

\subsection{Kemeny Score of Approx. Kemeny Rankings}

We can bound the difference of Kemeny scores for two Kemeny rankings of similar matrices.

\begin{lemma}[\appref{le: KS-approx-similar-matrices}]\label{le: KS-approx-similar-matrices}
Let $Q, \hat{Q} \in \winmatrix$ with $\|Q-\hat{Q}\|_1 \leq \epsilon$. Furthermore, let $\tau$ be a Kemeny ranking of $Q$ and $\hat{\tau}$ a Kemeny ranking of $\hat{Q}$. 
Then $\kscore(Q,\hat{\tau})-\kscore(Q,\tau) \leq \epsilon$.
\end{lemma}
\appendixproof{le: KS-approx-similar-matrices}{
\begin{proof}
 Because $Q, \hat{Q} \in \winmatrix$, we have for any arbitrary ranking $\succ$ on $[\arms]$ that
\begin{align*}
    &\|Q-\hat{Q}\|_1 = \sum_{(i,j)\in[\arms]\times[\arms]} |q_{ij} - \hat{q}_{ij}| \\
    &= \sum_{(i,j)\in \succ} |q_{ji} - \hat{q}_{ji}| +  |q_{ij} - \hat{q}_{ij}| \\
    &= \sum_{(i,j)\in \succ} |q_{ji} - \hat{q}_{ji}| + |(1-q_{ji}) - (1-\hat{q}_{ji})| \\
    &= \sum_{(i,j)\in \succ} |q_{ji} - \hat{q}_{ji}| + |\hat{q}_{ji} - q_{ji}| \\
    &= 2 \sum_{(i,j)\in \succ} |q_{ji} - \hat{q}_{ji}|.
\end{align*}
Here, the second equality holds because $q_{ii} = \hat{q}_{ii} = 0.5$ for all $i\in[\arms]$.
Since $\|Q-\hat{Q}\|_1 \leq \epsilon$, we also have $\sum_{(i,j)\in \succ} |q_{ji} - \hat{q}_{ji}| \leq \epsilon / 2$ for any ranking $\succ$.
Thus, for any ranking $\succ$,
\begin{align*}
    |\kscore(\hat{Q},\succ)-\kscore(Q,\succ)| &= |\sum_{(i,j)\in\succ} \hat{q}_{ji} - q_{ji}| \\
    &\leq \sum_{(i,j)\in\succ} |\hat{q}_{ji} - q_{ji}|\\
    &\leq \epsilon / 2.
\end{align*}
In particular, for $\tau$ and $\hat{\tau}$ we can derive the following bounds on $\kscore(\hat{Q},\hat{\tau})$ and $\kscore(Q,\tau)$:
\begin{align*}
    \kscore(\hat{Q},\tau) &\leq \kscore(Q,\tau) + \epsilon / 2\\
    \kscore(Q,\hat{\tau}) &\leq \kscore(\hat{Q},\hat{\tau}) + \epsilon / 2.
\end{align*}
Because $\hat{\tau}$ is a Kemeny ranking and minimise the Kemeny score w.r.t. $\hat{Q}$, we can conclude
$\kscore(Q,\hat{\tau}) \leq \kscore(\hat{Q},\hat{\tau}) + \epsilon / 2
    \leq \kscore(\hat{Q},\tau) + \epsilon / 2
    \leq \kscore(Q,\tau) + \epsilon$.
\end{proof}
}

Note that because $\tau$ is a Kemeny ranking w.r.t. $Q$, the difference of Kemeny scores of $\tau$ and $\hat{\tau}$ cannot be negative.
Lemma~\ref{le: KS-approx-similar-matrices} shows that with a close approximation of a matrix we can gain a Kemeny ranking that approximates the Kemeny score, i.e., minimal voter disagreement, well.

When sampling preferences, or more specifically feedback from dueling bandits, we can express the value of our approximation of the winning matrix by confidence intervals for matrix entries.
For this type of uncertainty quantification, we can adjust the approximation bounds from Lemma~\ref{le: KS-approx-similar-matrices}.

\begin{corollary}\label{cor:basic-conf-interval-approx-bound}
Let $Q, \hat{Q} \in \winmatrix$.
Consider Kemeny rankings $\tau$ of $Q$ and $\hat{\tau}$ of $\hat{Q}$. 
Suppose $q_{ij} \in [\hat{q}_{ij} - c_{ji}, \hat{q}_{ij} + c_{ij}]$ for some matrix $C \in \mathds{R}_{\geq0}^{\arms \times\arms}$ holds for all $i,j \in [\arms], i\neq j$ with probability $(1-\delta)$.
Then $\kscore(Q,\hat{\tau})-\kscore(Q,\tau) \leq 2\sum_{1 \leq i<j \leq\arms} \max(c_{ij},c_{ji})$ with probability $(1-\delta)$.
\end{corollary}

We can, in fact, refine this approximation bound for a Kemeny ranking $\hat{\tau}$ of $\hat{Q}+C$, with confidence bounds $C$.

\begin{lemma}[\appref{le: KS-approx-conf-bounds}]\label{le: KS-approx-conf-bounds}
Let $Q, \hat{Q} \in \winmatrix$ and $C \in \mathds{R}^{\arms \times\arms}$ be matrices such that $q_{ij} \in [\hat{q}_{ij} - c_{ji}, \hat{q}_{ij} + c_{ij}]$ holds for all $i,j \in [\arms], i\neq j$ with probability $(1-\delta)$.
Consider Kemeny rankings $\tau$ of $Q$ and $\hat{\tau}$ of $\hat{Q}+C$. 
Then $\kscore(Q,\hat{\tau})-\kscore(Q,\tau) \leq \sum_{1 \leq i<j \leq\arms} (c_{ij} + c_{ji})$ with probability $(1-\delta)$.
\end{lemma}
\appendixproof{le: KS-approx-conf-bounds}{
\begin{proof}
Let $Q, \hat{Q} \in \winmatrix$ and $C \in \mathds{R}^{\arms \times\arms}$ be such that $q_{ij} \in [\hat{q}_{ij} - c_{ji}, \hat{q}_{ij} + c_{ij}]$ holds for all $i,j \in [\arms], i\neq j$ with probability $(1-\delta)$.
Then, with probability $(1-\delta)$, we have
$q_{ji} \leq \hat{q}_{ji} + c_{ji}$ and $\hat{q}_{ji} \leq q_{ji} + c_{ij}$ for all $i,j \in [\arms], i\neq j$.
Thus, with probability $(1-\delta)$,
\begin{align*}
    \kscore(Q,\hat{\tau}) &= \sum\nolimits_{(i,j) \in \hat{\tau}} q_{ji} \\
    &\leq \sum\nolimits_{(i,j) \in \hat{\tau}} \hat{q}_{ji} + c_{ji}\\
   &= \kscore(\hat{Q}+C,\hat{\tau})\\
   &\leq \kscore(\hat{Q}+C,\tau)\\
   &= \sum\nolimits_{(i,j) \in \tau} \hat{q}_{ji} + c_{ji}\\
   &\leq \sum\nolimits_{(i,j) \in \tau} (q_{ji} + c_{ij}) + c_{ji}\\
   &= \kscore(Q,\tau) +  \sum\nolimits_{(i,j) \in \tau} (c_{ij} + c_{ji})\\
   &= \kscore(Q,\tau) +  \sum\nolimits_{1\leq i<j\leq \arms} (c_{ij} + c_{ji}).
\end{align*}
Here, the second inequality follows from $\hat{\tau}$ being a Kemeny ranking, and thus minimising the Kemeny score, w.r.t $\hat{Q}+C$. The last equality follows from $\tau$ being a complete order.
\end{proof}
}

Thus, to construct a good ranking in terms of Kemeny score, we can choose a Kemeny ranking w.r.t.  upper confidences of an approximated matrix of winning probabilities.

The above described approximation bounds for Kemeny scores are dependent on the quality of the approximation of winning probabilities for all pairs of arms. This is unsurprising, since the Kemeny score itself takes all winning probabilities into account with equal weight.
In the next section, we discuss concrete confidence bounds and PAC algorithms for Kemeny rankings w.r.t. Kemeny scores.

\section{PAC Algorithms for Kemeny Rankings w.r.t. Kemeny Scores}
\label{sec: PAC algos and sampling complexity}
\appendixsection{sec: PAC algos and sampling complexity}
To approximate a Kemeny ranking, 
we determine a sequence of pairs of arms to pull. 
We first consider strategies for the setting where we draw samples with replacement from Bernoulli distributions with the winning probabilities as means.
Thus, voters are drawn uniformly at random from an unknown population and we might ask the same voter the same query twice.
We then investigate the setting in which the population of voters is known and every voter may only be asked the same query once (sampling from a finite population without replacement).

\subsection{Sampling With Replacement}
Consider a matrix $Q \in \winmatrix$ of winning probabilities between arms such that every sample $(i,j)$ of two arms returns $i \succ j$ with probability $q_{ij}$.
Because confidence interval sizes are usually monotonically decreasing with the number of samples, Lemma~\ref{le: KS-approx-conf-bounds} suggests that for symmetric confidence bounds and without taking any further inference into account a uniform sampling strategy minimises the Kemeny-score-approximation error.
More concretely, it follows from the next proposition that Algorithm~\ref{alg: Kemeny_El_With_Replacement} is a PAC algorithm for approximating Kemeny rankings w.r.t. Kemeny scores.

\begin{algorithm}[hbt!]
 \caption{\\KemenyEl($Q, \rho, \delta$): Sampling with Replacement from  $Q\in \winmatrix$}
 \label{alg: Kemeny_El_With_Replacement}
\begin{algorithmic}
\STATE $\hat{Q} \gets \{1/2\}^{\arms \times \arms}$ \hspace{2.9cm} $\rhd$ Estimate of $Q$
\STATE $x:= \frac{\arms\cdot(\arms-1)}{\rho}$, $y := \ln{\frac{\delta}{\arms(\arms-1)}}$ \hspace{1.15cm} $\rhd$ Constants
\STATE $t:= \lceil-\frac{1}{2}x^2y\rceil, c:= \sqrt{-\frac{1}{2t}y}$ \hspace{1.05cm} $\rhd$ Sample Size and 
\STATE \hspace{5.3cm} Confidence Bound

\FOR{arms $i,j \in [\arms]$}
    \STATE $s_1, \dots s_t \sim \text{Bernoulli}(q_{ij})$
    \STATE $\hat{q}_{ij} \gets 1/t \sum_{i=1, \dots,t} s_i$
    \STATE $\hat{q}_{ji} \gets 1-\hat{q}_{ij}$
\ENDFOR

\STATE \textbf{return} $\kemeny(\hat{Q}+\mathds{1}\cdot c)$
\end{algorithmic}
\end{algorithm}

\begin{proposition}[\appref{pr: PAC-uniform-sampling-with-replacement}]\label{pr: PAC-uniform-sampling-with-replacement}
Let $\hat{Q} \in \winmatrix$ be the matrix of sample averages approximating some $Q \in \winmatrix$ after $t$ samples of each pair of arms, and let $c_\delta^t = \sqrt{\frac{1}{2t}y}$ with $y := -\ln{\frac{\delta}{\arms(\arms-1)}} > 1$.
Then, with probability $(1-\delta)$ we have $\kscore(Q,\hat{\tau})-\kscore(Q,\tau) \leq \arms\cdot(\arms-1)\cdot c_\delta^t$ for Kemeny ranking $\tau$ of $Q$ and Kemeny ranking $\hat{\tau}$ of $\hat{Q}+\mathds{1}\cdot c_\delta^t$.
\end{proposition}
\appendixproof{pr: PAC-uniform-sampling-with-replacement}{
\begin{proof}
Let $\tau$ be a Kemeny ranking of $Q$ and $\hat{\tau}$ a Kemeny ranking of $Q+I\cdot c_\delta^t$.
By Lemma~\ref{le: KS-approx-conf-bounds}, if $q_{ij} \in [\hat{q}_{ij} - c_\delta^t, \hat{q}_{ij} + c_\delta^t]$ holds for all $i,j \in [\arms], i\neq j$ with probability $(1-\delta)$, then  $\kscore(Q,\hat{\tau})-\kscore(Q,\tau) \leq \arms\cdot(\arms-1)\cdot c_\delta^t$ with probability $(1-\delta)$.
It thus remains to show that $q_{ij} \in [\hat{q}_{ij} - c_\delta^t, \hat{q}_{ij} + c_\delta^t]$ holds for all $i,j \in [\arms], i\neq j$ with probability $(1-\delta)$.

First, we can see that $q_{ij} \in [\hat{q}_{ij} - c_\delta^t, \hat{q}_{ij} + c_\delta^t]$, if and only if  $q_{ji} \in [\hat{q}_{ji} - c_\delta^t, \hat{q}_{ji} + c_\delta^t]$.
This is, because $q_{ij}  = 1- q_{ji}$ and $\hat{q}_{ij}  = 1- \hat{q}_{ji}$:
\begin{align*}
    & q_{ij} \in [\hat{q}_{ij} - c_\delta^t, \hat{q}_{ij} + c_\delta^t] \\
\Leftrightarrow 
    & q_{ij} \geq \hat{q}_{ij} - c_\delta^t \text{ and } \\  & q_{ij} \leq \hat{q}_{ij} + c_\delta^t \\
\Leftrightarrow 
    & 1 - q_{ij} \leq 1 - \hat{q}_{ij} + c_\delta^t \text{ and } \\  & 1 - q_{ij} \geq 1 - \hat{q}_{ij} - c_\delta^t \\
\Leftrightarrow 
    & q_{ji} \leq \hat{q}_{ji} + c_\delta^t \text{ and }  \\  & q_{ji} \geq \hat{q}_{ji} - c_\delta^t \\
\Leftrightarrow 
    & q_{ji} \in [\hat{q}_{ji} - c_\delta^t, \hat{q}_{ji} + c_\delta^t].
\end{align*}

By Hoeffding's inequality, see e.g.~\cite{Hoef94}, for every pair of arms $(i,j)$ we have that $\mathds{P}(|\hat{q}_{ij} - q_{ij}| \geq c_\delta^t) \leq 2 \cdot \exp(-2t(c_\delta^t)^2)$.

The probability that $q_{ij} \notin [\hat{q}_{ij} - c_\delta^t, \hat{q}_{ij} + c_\delta^t]$ for some pair of arms $(i,j)$ is 
\begin{align*}
    & \sum\nolimits_{1\leq i < j \leq \arms} \mathds{P}(|\hat{q}_{ij} - q_{ij}| \geq c_\delta^t) \\
    \leq & \sum\nolimits_{1\leq i < j \leq \arms} 2 \cdot \exp(-2t(c_\delta^t)^2)\\
    = & \arms (\arms - 1) \exp(-2t(c_\delta^t)^2)\\
    = & \arms (\arms - 1) \exp(\ln(\frac{\delta}{\arms(\arms-1)}))\\
    = & \delta.
\end{align*}
Thus, the probability that $q_{ij} \in [\hat{q}_{ij} - c_\delta^t, \hat{q}_{ij} + c_\delta^t]$ holds for all $i,j \in [\arms], i\neq j$ is $(1-\delta)$.
\end{proof}
}

Given a desired approximation rate $\rho$, we can reformulate Proposition~\ref{pr: PAC-uniform-sampling-with-replacement} and obtain the necessary number of samples.

\begin{corollary}\label{cor: sample-complexity-uniform-sampling}
For $\arms \geq 2$ arms, $Q\in \winmatrix$ with Kemeny ranking $\tau$, fixed approximation guarantee $0 < \rho \leq \frac{\arms(\arms-1)}{2}$ and probability $\delta$, Algorithm~\ref{alg: Kemeny_El_With_Replacement} is a $(\delta, \rho)$-PAC algorithm. That is, after $t < \frac{1}{2}x^2y$ samples of each pair of arms, Algorithm~\ref{alg: Kemeny_El_With_Replacement} returns ranking $\hat{\tau}$ such that with probability $(1-\delta)$, we have
 $\kscore(Q,\hat{\tau})-\kscore(Q,\tau) \leq \rho$. Here $x:= \frac{\arms\cdot(\arms-1)}{\rho}$ and $y := -\ln{\frac{\delta}{\arms(\arms-1)}} >1$.
\end{corollary}

\subsection{Sampling Without Replacement}
Suppose that, while sampling voters uniformly at random, 
none of the voters $v\in N$ can be asked twice about their preferences over a pair of arms.
Then the number $s$ out of $t$ asked voters that prefer $i$ to $j$ 
follows a hypergeometric distribution: $s \sim \text{Hypergeometric}(n, n\cdot q_{ij}, t)$.
In this case, we can refine the confidence bounds of our sampled estimation of the true fractions of voters $q_{ij}$ preferring $i$ to $j$.
This leads to better approximation for a fixed number of samples, or reversely, a lower sample complexity for a fixed approximation goal, compared to sampling with replacement.
More concretely, it follows from the next propositions that Algorithm~\ref{alg: Kemeny_El_Without_Replacement} is a PAC algorithm for Kemeny rankings w.r.t. Kemeny scores when sampling without replacement.

\begin{algorithm}[hbt!]
 \caption{\\KemenyEl($Q, \rho, \delta$): Sampling w/o Replacement from  $Q\in \prefmatrix$}
 \label{alg: Kemeny_El_Without_Replacement}
\begin{algorithmic}
\STATE $\hat{Q} \gets \{1/2\}^{\arms \times \arms}$ \hspace{3cm} $\rhd$ Estimate of $Q$
\STATE $x:= \frac{\arms\cdot(\arms-1)}{\rho}$, $y := -\ln{\frac{\delta}{\arms(\arms-1)}}$ \hspace{1.25cm} $\rhd$ Constants
\IF{$n<\frac{x^2y-4}{2}$} 
\STATE $t := \frac{-2n+x^2yn}{2n+x^2y}$, $c := \sqrt{\frac{(n-t)(t+1)}{2t^2n}y}$\\ \hspace{0.65cm}$\rhd$ Sample Size and Confidence Bound (small $n$)
\ELSE
\STATE $t := \frac{x^2y(n+1)}{2n+x^2y}$, $c := \sqrt{\frac{n-t+1}{2tn}y}$\\ \hspace{0.65cm}$\rhd$ Sample Size and Confidence Bound (large $n$)
\ENDIF

\FOR{arms $i,j \in [\arms]$}
    \STATE $s \sim \text{Hypergeometric}(n, n\cdot q_{ij}, t)$
    \STATE $\hat{q}_{ij} \gets s/t$
    \STATE $\hat{q}_{ji} \gets 1-\hat{q}_{ij}$
\ENDFOR

\STATE \textbf{return} $\kemeny(\hat{Q}+\mathds{1}\cdot c)$
\end{algorithmic}
\end{algorithm}

\begin{proposition}[\appref{pr: PAC-uniform-sampling-without-replacement}]\label{pr: PAC-uniform-sampling-without-replacement}
Consider $n >1$ voters with  preference matrix $Q \in \prefmatrix$ over $\arms \geq 2$ alternatives.
Let $\delta < 0.5$ be the approximation probability and $y := -\ln{\frac{\delta}{\arms(\arms-1)}} >1$.
Furthermore let $\hat{Q} \in \winmatrix$ be the matrix of sample averages after $t$ 
uniformly random samples from the population for each pair of arms (without replacement).

Then $c_{\delta,n}^t = \sqrt{\frac{n-t+1}{2tn}y}$ is a $(1-\delta)$-confidence bound for all entries in $\hat{Q}$. 
Furthermore, with probability $(1-\delta)$ we have $\kscore(Q,\hat{\tau})-\kscore(Q,\tau) \leq \arms\cdot(\arms-1)\cdot c_{\delta,n}^t$ for Kemeny ranking $\tau$ of $Q$ and Kemeny ranking $\hat{\tau}$ of $\hat{Q}+\mathds{1}\cdot c_{\delta,n}^t$.

For given $\rho$, we can guarantee an approximation $\kscore(Q,\hat{\tau})-\kscore(Q,\tau) \leq \rho$ with probability $(1-\delta)$ after 
$t = \frac{x^2y(n+1)}{2n + x^2y}$
samples (without replacement) of each pair of arms, where $x:= \frac{\arms\cdot(\arms-1)}{\rho}$.
\end{proposition}
\appendixproof{pr: PAC-uniform-sampling-without-replacement}{
\begin{proof}

Note that for the case of a fixed population and sampling without replacements, the number of samples (voters) that reveal $i \succ j$ follows a hypergeometric distribution. As a consequence, we can apply known tail bounds. 
By~\cite{Serf74} we know that 
\[\mathds{P}(\hat{q}_{ij} - q_{ij} \geq c_\delta^t) \leq  \exp(-2t(c_\delta^t)^2\frac{n}{n-t+1})\] for any $c_\delta^t > 0$ and for all pairs of arms $i,j \in [\arms]$.
By symmetry, we can retrieve the same bound for the other tail:
\begin{align*}
    &\mathds{P}(\hat{q}_{ij} - q_{ij} \leq - c_\delta^t)\\
    = &\mathds{P}(1 - \hat{q}_{ij} -1 + q_{ij} \geq c_\delta^t)\\
    = &\mathds{P}(\hat{q}_{ji} - q_{ji} \geq c_\delta^t)
\end{align*}
Thus, for all pairs of arms $i,j \in [\arms]$, \[\mathds{P}(|\hat{q}_{ij} - q_{ij}| \geq c_\delta^t) \leq  2\exp(-2t(c_\delta^t)^2\frac{n}{n-t+1}).\]

By our choice of $c_\delta^t$, we have that $c_\delta^t > 0$ as long as $t<n$. 
Furthermore, the probability that $q_{ij} \notin [\hat{q}_{ij} - c_\delta^t, \hat{q}_{ij} + c_\delta^t]$ for some pair of arms $(i,j)$ is 
\begin{align*}
    & \sum\nolimits_{1\leq i < j \leq \arms} \mathds{P}(|\hat{q}_{ij} - q_{ij}| \geq c_\delta^t) \\
    \leq & \sum\nolimits_{1\leq i < j \leq \arms} 2  \exp(-2t(c_\delta^t)^2\frac{n}{n-t+1})\\
    = & \arms (\arms - 1) \exp(-2t(c_\delta^t)^2\frac{n}{n-t+1})\\
    = & \arms (\arms - 1) \exp(\ln(\frac{\delta}{\arms(\arms-1)}))\\
    = & \delta.
\end{align*}
Thus, the probability that $q_{ij} \in [\hat{q}_{ij} - c_\delta^t, \hat{q}_{ij} + c_\delta^t]$ holds for all $i,j \in [\arms], i\neq j$ is $(1-\delta)$.

Now assume a fixed approximation guarantee $\rho$ is given, and that $t=\frac{x^2y(n+1)}{2n + x^2y}$.
Then we have
\begin{align*}
    t &= \frac{x^2y(n+1)}{2n + x^2y} \\
    \Leftrightarrow 
    2tn &= x^2y(n-t+1)\\
    \Leftrightarrow 
    2tn &= (\frac{\arms\cdot(\arms-1)}{\rho})^2y(n-t+1) \\
    \Leftrightarrow 
    \rho &= \arms\cdot(\arms-1)\cdot \sqrt{\frac{n-t+1}{2tn}y} \\
    \Leftrightarrow 
    \rho &= \arms\cdot(\arms-1)\cdot c_\delta^t .
\end{align*}
Thus with probability $(1-\delta)$, we have $\kscore(Q,\hat{\tau})-\kscore(Q,\tau) \leq \rho$ for $t=\frac{x^2y(n+1)}{2n + x^2y}$ samples (without replacement) of each pair of arms. 
\end{proof}
}

For the case that we have more samples than half of the population size, i.e., $t > n/2$, and a limited population size, we can apply better bounds on confidences over the winning probabilities. This leads to the following result.

\begin{proposition}[\appref{pr: PAC-uniform-sampling-without-replacement-large-sample-sizes}]\label{pr: PAC-uniform-sampling-without-replacement-large-sample-sizes}
Assume a fixed voter population of size $n >1$ with preference matrix $Q \in \prefmatrix$ over $\arms \geq 2$ alternatives.
Let $\delta < 0.5$ be the approximation probability and $y := -\ln{\frac{\delta}{\arms(\arms-1)}} >1$.
Furthermore, let $\hat{Q} \in \winmatrix$ be the matrix of sample averages after $t>n/2$ uniformly random samples 
for each pair of arms (without replacement).

Then $c_{\delta,n}^t = \sqrt{\frac{(n-t)(t+1)}{2t^2n}y}$ is a $(1-\delta)$-confidence bound for all entries in $\hat{Q}$.
Furthermore, with probability $(1-\delta)$ we have $\kscore(Q,\hat{\tau})-\kscore(Q,\tau) \leq \arms\cdot(\arms-1)\cdot c_{\delta,n}^t$ for Kemeny ranking $\tau$ of $Q$ and Kemeny ranking $\hat{\tau}$ of $\hat{Q}+\mathds{1}\cdot c_{\delta,n}^t$.

For given $\rho$ and if the population is of size $n<\frac{x^2y-4}{2}$, we can guarantee an approximation  $\kscore(Q,\hat{\tau})-\kscore(Q,\tau) \leq \rho$ with probability $(1-\delta)$ after 
$t=\frac{x^2yn + 2n}{2n + x^2y}$
samples (without replacement) of each pair of arms, where $x:= \frac{\arms\cdot(\arms-1)}{\rho}$.
\end{proposition}
\appendixproof{pr: PAC-uniform-sampling-without-replacement-large-sample-sizes}{
\begin{proof}
As argued in the proof of Proposition~\ref{pr: PAC-uniform-sampling-without-replacement}, for all pairs of arms $i,j \in [\arms]$, \[\mathds{P}(|\hat{q}_{ij} - q_{ij}| \geq c) \leq  2\exp(-2t(c)^2\frac{n}{n-t+1}).\]
By symmetry of sampled voters and remaining voters, we can achieve a tightened bound.
For this purpose, let $\hat{q}_{ij}^t$ denote the sample average after taking $t$ samples without replacement and $\hat{q}_{ij}^{n-t} = \frac{nq_{ij} - t\hat{q}_{ij}^t}{n-t}$ the average of the remaining unsampled voters. Then $\hat{q}_{ij}^t = \frac{nq_{ij} - (n-t)\hat{q}_{ij}^{n-t}}{t}$ and
\begin{align*}
&\mathds{P}(|\hat{q}_{ij}^t - q_{ij}| \geq c) \\
= & \mathds{P}(|\frac{nq_{ij} - (n-t)\hat{q}_{ij}^{n-t}}{t} - q_{ij}| \geq c) \\
= & \mathds{P}(|\frac{(n-t)q_{ij} - (n-t)\hat{q}_{ij}^{n-t}}{t}| \geq c) \\
= & \mathds{P}(|q_{ij} - \hat{q}_{ij}^{n-t}| \geq \frac{t}{n-t}c) \\
= & \mathds{P}(|\hat{q}_{ij}^{n-t} - q_{ij}| \geq \frac{t}{n-t}c) \\
\leq & 2\exp(-2(n-t)(\frac{t}{n-t}c)^2\frac{n}{n-(n-t)+1}) \\
= & 2\exp(-2tc^2\frac{nt}{(n-t)(t+1)}).
\end{align*}
To see that this bound is actually tighter for $t>n/2$ we compare:
\begin{align*}
    & 2\exp(-2t(c)^2\frac{n}{n-t+1}) && > 2\exp(-2tc^2\frac{nt}{(n-t)(t+1)}) \\
    \Leftrightarrow 
    & \frac{n}{n-t+1} && < \frac{nt}{(n-t)(t+1)} \\
    \Leftrightarrow 
    & (n-t)(t+1) && < nt-t^2+t \\
    \Leftrightarrow 
    & nt + n - t^2 -t && < nt-t^2+t \\
    \Leftrightarrow 
    &  n/2 && < t 
\end{align*}
By our choice of $c_{\delta,n}^t = \sqrt{\frac{(n-t)(t+1)}{2t^2n}y}$, we have that the probability that $q_{ij} \notin [\hat{q}_{ij} - c_\delta^t, \hat{q}_{ij} + c_\delta^t]$ for some pair of arms $(i,j)$ is 
\begin{align*}
    & \sum\nolimits_{1\leq i < j \leq \arms} \mathds{P}(|\hat{q}_{ij} - q_{ij}| \geq c_\delta^t) \\
    \leq & \sum\nolimits_{1\leq i < j \leq \arms} 2  \exp(-2t(c_\delta^t)^2\frac{nt}{(n-t)(t+1)})\\
    = & \arms (\arms - 1) \exp(-2t(c_\delta^t)^2\frac{nt}{(n-t)(t+1)})\\
    = & \arms (\arms - 1) \exp(\ln(\frac{\delta}{\arms(\arms-1)}))\\
    = & \delta.
\end{align*}
Thus, the probability that $q_{ij} \in [\hat{q}_{ij} - c_\delta^t, \hat{q}_{ij} + c_\delta^t]$ holds for all $i,j \in [\arms], i\neq j$ is $(1-\delta)$.

Now assume a fixed approximation guarantee $\rho$ is given, and that $t=\frac{x^2yn + 2n}{2n + x^2y}$ and $n<\frac{x^2y-4}{2}$.
Then we have
\begin{align*}
    t &= \frac{x^2yn + 2n}{2n + x^2y} \\
    \Leftrightarrow 
    (2n + x^2y)t &= x^2yn + 2n \\
    \Leftrightarrow 
    2n(t-1)  &= x^2y(n-t) \\
    \Leftrightarrow 
    2n(t-1)  &= (\frac{\arms\cdot(\arms-1)}{\rho})^2y(n-t) \\
    \Leftrightarrow 
    \rho^2  &= \arms^2\cdot(\arms-1)^2\frac{(n-t)}{2n(t-1)}y \\
    \Leftrightarrow 
    \rho  &= \arms\cdot(\arms-1)\sqrt{\frac{(n-t)}{2n(t-1)}y} \\
    \Rightarrow 
    \rho &> \arms\cdot(\arms-1)\cdot \sqrt{\frac{(n-t)(t+1)}{2nt^2}y} \\
    \Leftrightarrow 
    \rho &> \arms\cdot(\arms-1)\cdot c_\delta^t .
\end{align*}

Furthermore,
\begin{align*}
    n &<\frac{x^2y+4}{2} \\
    \Leftrightarrow 
    x^2yn + 4n & > 2n^2 \\
    \Leftrightarrow 
    2x^2yn + 4n & > 2n^2 + x^2yn \\
    \Leftrightarrow 
    x^2yn + 2n & > n/2(2n + x^2y) \\
    \Leftrightarrow 
    \frac{x^2yn + 2n}{2n + x^2y} & > n/2 \\
    \Leftrightarrow 
    t & > n/2
\end{align*}

We can now apply our previous results for confidence and approximation bounds.
Thus with probability $(1-\delta)$, we have $\kscore(Q,\hat{\tau})-\kscore(Q,\tau) \leq \rho$ for $t=\frac{x^2yn + 2n}{2n + x^2y}$ samples (without replacement) of each pair of arms. 
\end{proof}
}

\subsection{Comparison}
Note that Proposition~\ref{pr: PAC-uniform-sampling-without-replacement} and \ref{pr: PAC-uniform-sampling-without-replacement-large-sample-sizes} use improved confidence bounds for the estimated winning probabilities compared to Proposition~\ref{pr: PAC-uniform-sampling-with-replacement}. This is not surprising, since we are sampling from a fixed population without replacement, and expect the confidence bound to be dependent on the relation between the number of samples and the population size.
Furthermore, unsurprisingly, this improves the sample complexity. 
We can quantify the advantage as follows.

\begin{corollary}\label{cor: summary-conf-bouds-sample-complexity}

Let $x:= \frac{\arms\cdot(\arms-1)}{\rho}$ and $y := -\ln{\frac{\delta}{\arms(\arms-1)}} >1$.

Consider the confidence bounds after $t$ samples from Proposition~\ref{pr: PAC-uniform-sampling-with-replacement}, Proposition~\ref{pr: PAC-uniform-sampling-without-replacement} and Proposition~\ref{pr: PAC-uniform-sampling-without-replacement-large-sample-sizes}:
\[c = \sqrt{\frac{1}{2t}y}, c' = \sqrt{\frac{n-t+1}{2tn}y}, c'' = \sqrt{\frac{(n-t)(t+1)}{2t^2n}y}.\]
Then we have $c' = \sqrt{\frac{n-t+1}{n}}c$ and $c'' = \sqrt{\frac{(n-t)(t+1)}{tn}}c$, i.e., tighter confidence bounds when sampling without replacement. Furthermore, $c'> c''$ if and only if $t > n/2$.

For fixed approximation guarantee $\frac{\arms(\arms-1)}{n} \geq \rho > 0$, consider the sample complexities from Proposition~\ref{pr: PAC-uniform-sampling-with-replacement},~\ref{pr: PAC-uniform-sampling-without-replacement},~\ref{pr: PAC-uniform-sampling-without-replacement-large-sample-sizes}:
\[t=\frac{1}{2}x^2y, t' = \frac{x^2y(n+1)}{2n + x^2y}, t'' = \frac{x^2yn +2n}{2n + x^2y}.\]
Then we have $t' = \frac{2(n+1)}{2n + x^2y}t$ and $t'' = \frac{2n +4n/(x^2y)}{2n + x^2y}t$.
For the reasonable assumption that $2/n < \rho$, $\arms\geq 2$ and $n\geq 2$ this means that fewer samples are needed when sampling without replacement than with replacement.
Furthermore, for $n<\frac{x^2y-4}{2}$ we have $t' > t''$.
\end{corollary}

This comparison also shows that the more samples are taken, the starker the difference between confidence bounds when sampling with and without replacement.
Furthermore, in order to reach an approximation of $\rho$, the smaller $\rho$ and $\delta$, the fewer samples we need when sampling without replacement compared to sampling with replacement. 

\section{Pruning of Confidence Intervals} \label{sec: pruning confidence intervals}
\appendixsection{sec: pruning confidence intervals}
In Section~\ref{sec: PAC algos and sampling complexity} we estimated the confidence intervals of the winning probabilities $q_{ij}$ by the use of some known concentration inequalities.
This treatment delivered symmetric bounds that, given the same number of samples, are the same for all pairs of arms $i,j \in [\arms]$.
However, the values $q_{ij}$ of a preference matrix $Q \in \prefmatrix$ are not independent. In fact, Lemma~\ref{le: cond-for-pref-matrix} implies that  $q_{ij} + q_{ji} = 1$,  for all $i,j \in [\arms]$.
We can restrict confidence intervals to comply with this constraint. 

\begin{lemma}[\appref{le: pruning-by-symmetry}]\label{le: pruning-by-symmetry}
Assume that the sampled means $\hat{q}_{ij}$ after some samples from a preference matrix $Q\in\prefmatrix$ have (possibly asymmetric) confidence intervals: \[q_{ij} \in [\hat{q}_{ij} - \underline{c}_{ij}, \hat{q}_{ij} + \overline{c}_{ij}]\] for all arms $i,j \in [\arms]$ with probability $(1-\delta)$.
Define matrix $C\in [0,1]^{\arms\times\arms}$ by $c_{ij} := \min(\overline{c}_{ij},\underline{c}_{ji})$.
Then with probability $(1-\delta)$ for all arms $i,j \in [\arms]$:
\begin{align} \label{eq: symmetric confidence bounds}
    q_{ij} \in [\hat{q}_{ij} - c_{ji}, \hat{q}_{ij} + c_{ij}].
\end{align} 
\end{lemma}
\appendixproof{le: pruning-by-symmetry}{
\begin{proof}
Because $q_{ij} + q_{ji} = 1$ and $\hat{q}_{ij} + \hat{q}_{ji} = 1$, we have $q_{ij} = 1 -  q_{ji} \geq 1 - \hat{q}_{ji} - \overline{c}_{ji} \geq \hat{q}_{ij} - \overline{c}_{ji}$ for all arms $i,j \in [\arms]$ with probability $(1-\delta)$.
Similarly, $q_{ij} \leq \hat{q}_{ij} + \underline{c}_{ji}$ for all arms $i,j \in [\arms]$ with probability $(1-\delta)$.

Thus, for $c_{ij} := \min(\overline{c}_{ij},\underline{c}_{ji})$, we can refine the confidence bounds as
\begin{align*}
    & q_{ij} \in [\max(\hat{q}_{ij} - \underline{c}_{ij}, \hat{q}_{ij} - \overline{c}_{ji}), &&\min(\hat{q}_{ij} + \overline{c}_{ij}, \hat{q}_{ij} + \underline{c}_{ji})] \\
    \Leftrightarrow
    & q_{ij} \in [\hat{q}_{ij} - \min(\underline{c}_{ij}, \overline{c}_{ji}), &&\hat{q}_{ij} + \min(\overline{c}_{ij}, \underline{c}_{ji})] \\
    \Leftrightarrow
    & q_{ij} \in [\hat{q}_{ij} - c_{ji}, &&\hat{q}_{ij} + c_{ij}]     
\end{align*} 
for all arms $i,j \in [\arms]$ with probability $(1-\delta)$.
\end{proof}
}

Lemma~\ref{le: cond-for-pref-matrix} also implies that for a preference matrix $Q\in\prefmatrix$ a triangle inequality must hold for all triples of arms: $q_{lj} + q_{ji}\geq q_{li}$ for all $i,j,l \in [\arms]$.
We can comply with this constraint by pruning confidence intervals further.

\begin{lemma}[\appref{le: pruning-by-triangle-ineq}]\label{le: pruning-by-triangle-ineq}
Assume that the sampled means $\hat{q}_{ij}$ after some samples from a preference matrix $Q\in\prefmatrix$ have (possibly asymmetric) confidence intervals: \[q_{ij} \in [\hat{q}_{ij} - c_{ji}, \hat{q}_{ij} + c_{ij}]\] for all arms $i,j \in [\arms]$ with probability $(1-\delta)$.
Define matrix $\dot{C}\in [0,1]^{\arms\times\arms}$ by \[\dot{c}_{ij} := \min{(c_{ij}, \min_{l\in[\arms]} \hat{q}_{il} + c_{il} + \hat{q}_{lj} + c_{lj} - \hat{q}_{ij})}.\]
Then with probability $(1-\delta)$ for all arms $i,j \in [\arms]$:
\begin{align} \label{eq: triangle ineq confidence bounds}
    q_{ij} \in [\hat{q}_{ij} - \dot{c}_{ji}, \hat{q}_{ij} + \dot{c}_{ij}].
\end{align} 
\end{lemma}
\appendixproof{le: pruning-by-triangle-ineq}{
\begin{proof}
    By Lemma~\ref{le: cond-for-pref-matrix}, $q_{il} + q_{lj} \geq q_{ij}$ for any $i,j,l \in [\arms]$. 
    For 
    $q_{il} \in [\hat{q}_{il} - c_{li}, \hat{q}_{il} + c_{il}]$
    and 
    $q_{lj} \in [\hat{q}_{lj} - c_{jl}, \hat{q}_{lj} + c_{lj}]$,
    we have 
    $\hat{q}_{il} + c_{il} + \hat{q}_{lj} + c_{lj} \geq q_{ij}.$
Thus, $\dot{c}_{ij} := \min{(c_{ij}, \min_{l\in[\arms]} \hat{q}_{il} + c_{il} + \hat{q}_{lj} + c_{lj} - \hat{q}_{ij})}$
is an upper $(1-\delta)$-confidence bound on $q_{ij}$.
To see that this is also a lower $(1-\delta)$-confidence bound on $q_{ji}$, consider the following:
\begin{align*}
    &\hat{q}_{il} + c_{il} + \hat{q}_{lj} + c_{lj} &&\geq  q_{ij} \\
    \Leftrightarrow
    &1-\hat{q}_{il} - c_{il} - \hat{q}_{lj} - c_{lj} &&\leq 1- q_{ij} \\
    \Leftrightarrow
    &1-\hat{q}_{il} - c_{il} - \hat{q}_{lj} - c_{lj} &&\leq q_{ji} .
\end{align*}
Hence, for all arms $i,j,l \in [\arms]$ with probability $(1-\delta)$
\begin{align*}
&\max(\hat{q}_{ji} - c_{ij}, 1-\hat{q}_{il} - c_{il} - \hat{q}_{lj} - c_{lj}) &&\leq q_{ji}\\
\Leftrightarrow
&\hat{q}_{ji} - \min(c_{ij}, -1 + \hat{q}_{il} + c_{il} + \hat{q}_{lj} + c_{lj} + \hat{q}_{ji}) &&\leq q_{ji}\\
\Leftrightarrow
&\hat{q}_{ji} - \min(c_{ij}, \hat{q}_{il} + c_{il} + \hat{q}_{lj} + c_{lj} - \hat{q}_{ij}) &&\leq q_{ji}.
\end{align*}
Because the inequalities hold for any triple $i,j,l \in [\arms]$, we have shown that $q_{ij} \in [\hat{q}_{ij} - \dot{c}_{ji}, \hat{q}_{ij} + \dot{c}_{ij}]$
for all arms $i,j \in [\arms]$ with probability $(1-\delta)$.
\end{proof}
}

For better understanding, we include an example of pruning confidence intervals in Appendix-Section~\ref{subsec: example of pruning}. \\
After pruning intervals 
once,
it might be feasible to prune the new confidence intervals even more. 
This procedure is described by Algorithm~\ref{alg: Kemeny_El_With_Pruning}.
Note that Lemma~\ref{le: pruning-by-triangle-ineq} allows $\dot{c}_{ij}$ to be negative, indicating that the sample mean $\hat{q}_{ij}$ overestimates the actual winning probability $q_{ij}$.
This does not influence the approximation result given in Lemma~\ref{le: KS-approx-conf-bounds}. However, we require that $1 \geq \dot{c}_{ij} + \hat{q}_{ij} \geq 0$ which is expressed in Algorithm~\ref{alg: Kemeny_El_With_Pruning} in the 4th and 10th line.
Algorithm~\ref{alg: Kemeny_El_With_Pruning} converges, because in every iteration of the repeat-loop at least one value $\dot{c}_{ij}$ strictly decreases and values never increase while maintaining a lower bound of $-\hat{q}_{ij}$. Furthermore, the values of initial confidence bounds and sample averages build a discrete set of values such that there are countably many combinations that can build the minimum for the updates.

We can now refine the elicitation methods from Section~\ref{sec: PAC algos and sampling complexity}.

\toappendix{
\subsection{An Example for Pruning Confidence Intervals}\label{subsec: example of pruning}
    Consider the following preference profile $P$ with preference matrix $Q$.
    \[ P = \{100: 1 \succ 2 \succ 3, 50: 1 \succ 3 \succ 2, 50: 2 \succ 3 \succ 1\},\] 
    \[ Q =   \begin{pmatrix}
        0.5  & 0.75 & 0.75 \\
        0.25 & 0.5  & 0.75 \\
        0.25 & 0.25 & 0.5
        \end{pmatrix}, 
    \]
    Let $\hat{Q}$ approximate $Q$ with upper and lower confidence bounds $\overline{C}$ and $\underline{C}$, respectively, given by:
    \[\hat{Q} =   \begin{pmatrix}
            0.5 & 0.9 & 0.6 \\
            0.1 & 0.5 & 0.9 \\
            0.4 & 0.1 & 0.5
            \end{pmatrix}, \]
    \[\underline{C} =   \begin{pmatrix}
            0.0 & 0.2  & 0.2 \\
            0.1 & 0.0  & 0.2 \\
            0.2 & 0.1  & 0.0
            \end{pmatrix}, 
    \overline{C} =   \begin{pmatrix}
            0.0 & 0.25 & 0.2 \\
            0.15 & 0.0 & 0.25 \\
            0.2 & 0.15 & 0.0
            \end{pmatrix}. 
    \]
    By Lemma~\ref{le: pruning-by-symmetry}, we can refine the confidence bounds to $C$, and by Lemma~\ref{le: pruning-by-triangle-ineq} further to $\dot{C}$:
    \[C =   \begin{pmatrix}
            0.0 & 0.1 & 0.2 \\
            0.15 & 0.0 & 0.1 \\
            0.2 & 0.15 & 0.0
            \end{pmatrix},
    \dot{C}=   \begin{pmatrix}
            0.0 & 0.1 & 0.2 \\
            0.15 & 0.0 & 0.1 \\
            0.1 & 0.15 & 0.0
            \end{pmatrix}.\]
    This is because 
    \begin{align*}
        \dot{c}_{31} := & \min{(c_{31}, \hat{q}_{32} + c_{32} + \hat{q}_{21} + c_{21} - \hat{q}_{31})}  \\
        = &\min(0.2, 0.1) = 0.1.
    \end{align*}
    These bounds are also not unexpected since, e.g., $\hat{q}_{12} + \overline{c_{12}} = 1.15$ while $q_{12} \leq 1$. 
}

\begin{algorithm}[hbt!]
 \caption{PruningCI($\hat{Q}, \underline{C}, \overline{C}$)\\ \footnotesize{Pruning Confidence Intervals $q_{ij} \in [\hat{q}_{ij} - \underline{c}_{ij}, \hat{q}_{ij} + \overline{c}_{ij}]$ for all $i,j\in [\arms]$ of preference matrix $Q\in \prefmatrix$}}
 \label{alg: Kemeny_El_With_Pruning}
\begin{algorithmic}
\STATE $c_{ij} \gets \min(\overline{c_{ij}}, \underline{c_{ji}})$   for all $i,j\in [\arms]$\\ 
$\rhd$ Upper/Lower Confidence Bounds 
based on Lemma~\ref{le: pruning-by-symmetry} \\
\STATE $c_{ij} \gets \min(c_{ij}, 1 - \hat{q}_{ij})$  for all $i,j\in [\arms]$\\ 
$\rhd$ Upper/Lower Confidence Bounds based on $q_{ij} \in [0,1]$ \\
\STATE $\dot{C} \gets C$
\REPEAT
    \STATE $C \gets \dot{C}$
    \FOR{arms $i,j \in [\arms]$}
        \STATE $\dot{c}_{ij} \gets \min{(c_{ij}, \min \limits_{l\in[\arms], l\neq i,j} \hat{q}_{il} + c_{il} + \hat{q}_{lj} + c_{lj} - \hat{q}_{ij})}$
        \STATE $\dot{c}_{ij} \gets \max{(-\hat{q}_{ij}, \dot{c}_{ij})}$
    \ENDFOR
\UNTIL{$C==\dot{C}$}

\STATE \textbf{return} $\dot{C}$
\end{algorithmic}
\end{algorithm}

\subsection{Adaptive Sampling Strategies}
Recall that our approximation bounds for Kemeny rankings from Section~\ref{sec: kemenys rule} are determined by the confidence intervals' sizes. 
Furthermore, the confidence bounds for sampling with or without replacement given in Section~\ref{sec: PAC algos and sampling complexity} can only be influenced by the number of pulls of the pairs of arms.
Thus uniform sampling guarantees to always sample from a pair of arms with the largest confidence interval.
However, when pruning confidence intervals after every new sample, the "most uncertain" pair of arms, i.e., one with the largest confidence interval, might not necessarily be one with the lowest number of pulls so far.
We can also estimate the outcome and effect of the next $\ell$ samples.
To simplify notation and computational complexity, we concentrate on such look-aheads with $l = 1$. 
This allows the following sample strategies:
\begin{itemize}
    \item \textbf{Uniform}: Sample a pair with minimal number of pulls.
    \item \textbf{Opportunistic}: Sample a pair with maximal confidence interval size.
    \item \textbf{Optimistic}: Sample a pair which, for the best case outcome, gives a maximal reduction in the sum of all confidence interval sizes.
    \item \textbf{Pessimistic}: Sample a pair which, for the worst case outcome, gives a maximal reduction in the sum of all confidence interval sizes.
    \item \textbf{Bayesian/Realistic}: Sample a pair which, in expectation w.r.t. the current estimation of winning probabilities, gives a maximal reduction in the sum of all confidence interval sizes.
\end{itemize}

Note that if in one round no pruning is possible for any sample outcome for any pair of arms, all sampling strategies collapse to opportunistic sampling.
If additionally all pairs have been pulled equally often, this is the same as uniform sampling as any sample outcome will give same reduction.

\subsection{Experiments}

\begin{figure*}[t]
    \centering
     \begin{subfigure}[b]{0.49\textwidth}
         \centering
         \includegraphics[trim={0 0 0 1.5cm},clip,width=\textwidth]{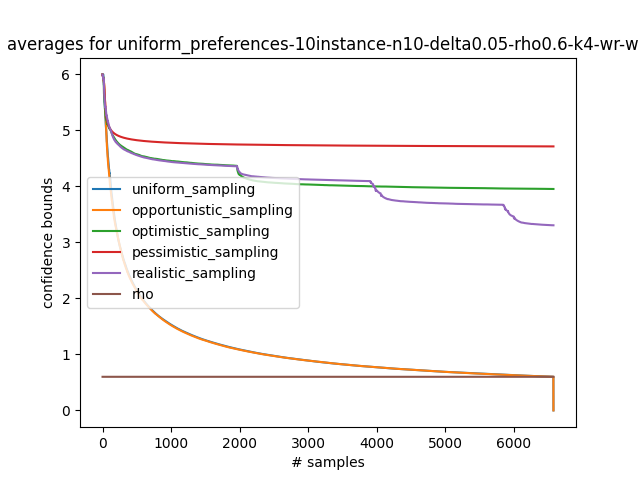}
         \caption{Sampling With Replacement. Averages over $10$ instances. \\$\arms=4$, $\rho = 0.6$. Average true Kemeny score $2.08$.}
         \label{fig:sample-with-repl}
     \end{subfigure}
     \hfill
     \begin{subfigure}[b]{0.49\textwidth}
         \centering
         \includegraphics[trim={0 0 0 1.5cm},clip,width=\textwidth]{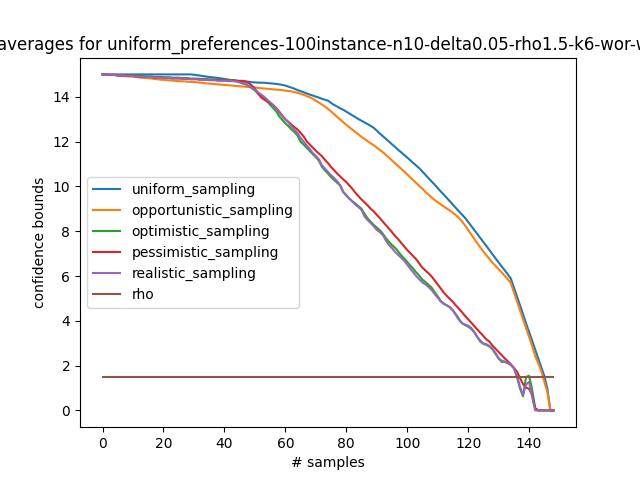}
         \caption{Sampling Without Replacement. Averages over $100$ instances. \\$\arms=6$, $\rho = 1.5$. Average true Kemeny score $5.618$.}
         \label{fig:sample-without-repl}
     \end{subfigure}
    \caption{Average Confidence Bounds for uniformly at random generated instances with $n=10$ voters. $\delta = 0.05$.}
    \label{fig: plots-comparison-sample-with-pruning-k=7}
\end{figure*}
\paragraph{Setup:} We generate preference matrices $Q \in \prefmatrix$ for $n = 10$ voters uniformly at random for up to $\arms = 9$ arms.
We set an approximation value of $\rho = 0.1 \frac{\arms (\arms-1)}{2}$, i.e., $10\%$ of the worst case difference in approximated and optimal Kemeny score, and approximation probability $(1-\delta) = 0.95$.
For better comparison, we compare all sampling methods on the same preference matrices. Furthermore, we average all our results over runs on several instances.
Note that the computational complexity for sampling with replacements only allows us to conduct experiments on 10 instances and for few arms, while for sampling without replacement we use 100 instances for experiments with up to $9$ arms. To compare Kemeny scores at every sampling step, it is required to compute Kemeny rankings which is a known NP-hard problem. We use a simple ILP formulation for this. All code is written in Python and is publicly available under \url{https://github.com/annemage/Eliciting-Kemeny-Rankings}.

\paragraph{Results and Analysis:}
As all sampling methods correspond to uniform sampling when no pruning is taken into account, we restrict our attention here to the case where we apply Algorithm~\ref{alg: Kemeny_El_With_Pruning} to the confidence intervals indicated in Corollary ~\ref{cor: summary-conf-bouds-sample-complexity} w.r.t. current numbers of pulls for every new sample.
Unsurprisingly, our experiments show that the Kemeny score for a current approximation is closer to the true Kemeny score than the approximation bounds given by the total length of confidence intervals (see Lemma~\ref{le: KS-approx-conf-bounds}) would suggest, i.e., our approximation bounds are not tight.
Because it is the approximation bounds, however, that influence the sample strategies, we show a comparison of these in
Figure ~\ref{fig: plots-comparison-sample-with-pruning-k=7} 
for a sequence of samples.
Note that here the length of the x-axis corresponds to the theoretical sample complexity given in~Corollary ~\ref{cor: summary-conf-bouds-sample-complexity}. 
Obviously, the two plots confirm that sampling without replacement is much more efficient for all sample methods.
Note that if for an instance sampling terminated because the approximation bound $\rho$ was reached early, averages are taken over the remaining instances only.

It can be easily seen in Figure ~\ref{fig:sample-with-repl} that when sampling with replacement only uniform and opportunistic sampling reach the given approximation bound $\rho$ within the theoretical sample complexity of $6576$ samples.
Note that in these experiments the look-ahead methods (optimistic, pessimistic and realistic sampling) often get stuck by sampling the same pairs of arms which apparently do not lead to enough pruning of confidence intervals to be worth it.
This is surprising especially for pessimistic sampling, as this method should be conservative w.r.t. pruning possibilities.
Furthermore, uniform and opportunistic sampling behave exactly the same and reach the desired approximation $\rho$ in exactly the theoretical sample complexity. This might imply that pruning has no effect in these instances which could be an artifact of how the confidence intervals are computed in this case.

On the other hand, it can easily be seen from~Figure ~\ref{fig:sample-without-repl} that for sampling without replacement even uniform sampling needs in average less than the theoretical number of samples
indicated in Corollary ~\ref{cor: summary-conf-bouds-sample-complexity} when pruning is applied, showing that pruning of confidence intervals is applicable in our instances. 
Here, 
the adaptive sampling strategies show a clear advantage over sampling uniformly. 
While opportunistic sampling seems the best choice when only sampling few pairs of arms, it is outperformed by optimistic, pessimistic and realistic (Bayesian) look-aheads for more samples. 
Naturally, the look-ahead methods can anticipate any possibility of pruning the confidence intervals, which seems to give them an advantage in the long run.
However, at this point we would like to point out that opportunistic sampling has much lower computational complexity than the look-ahead methods, which is an obvious advantage in practice.
Optimistic and realistic sampling behave similarly. They need fewer samples and give a better approximation than pessimistic sampling.
This might be because pessimistic sampling  is too conservative in reacting to chances of pruning.

For other numbers of arms the experimental results show similar behaviour.
Please refer to Section~\ref{subsec: additional-experimental results} for more experimental results regarding other numbers of arms and a comparison to preference profiles generated from a Mallows model~\cite{Mall57} with $\phi = 0.2$ (i.e., preferences being fairly similar) or uniformly at random generated single-peaked preferences which always admit a Condorcet winner.

\toappendix
{
\subsection{Additional Implementation Details \& Experimental Results}
\label{subsec: additional-experimental results}

\paragraph{Implementation:}
Our experiments / algorithms are implemented in Python version 3.9 using: numpy (np), pandas, itertools, math, copy, random, matplotlib, and ortools (for exact computation of Kemeny rankings via an integer linear programming formulation).
For reproducability, we set random seeds random.seed(24) and
np.random.seed(24).
All experiments were run on a Windows machine with Intel(R) Core(TM) i7-10510U CPU and 16.0 GB of RAM.

Note that we round any confidence bounds to 5 digits to avoid numerical issues.
Furthermore, sampling with replacement for larger numbers of arms requires many pulls making experiments computationally costly.
Furthermore, determining the actual distance of Kemeny scores between an approximate Kemeny ranking and the true Kemeny ranking at any sampling step is immensely costly, since computing Kemeny rankings is NP-hard.
We thus resort to only averaging over 10 instead of 100 randomly generated instances for some sizes of arms, and furthermore might only report the approximation bounds (i.e. total length of confidence bounds) and not the true approximation values.

\paragraph{Uniformly at Random Sampled Instances}
We give additional plots for sampling from instances that were generated uniformly at random in the following. These exhibit similar behaviour as discussed in the main paper.

\textbf{Pruning of confidence intervals after every sample.}\\
Sampling without replacement: Figure~\ref{fig:uar-wor-wp1},~\ref{fig:uar-wor-wp2},~\ref{fig:uar-wor-wp3},~\ref{fig:uar-wor-wp4},~\ref{fig:uar-wor-wp5},~\ref{fig:uar-wor-wp6},~\ref{fig:uar-wor-wp7}.\\
Sampling with replacement: Figure ~\ref{fig:uar-wr-wp1},~\ref{fig:uar-wr-wp2},~\ref{fig:uar-wr-wp3},~\ref{fig:uar-wr-wp4},~\ref{fig:uar-wr-wp5}.

\textbf{No pruning applied.}\\
Sampling without replacement:  Figure ~\ref{fig:uar-wor-wop1},~\ref{fig:uar-wor-wop2},~\ref{fig:uar-wor-wop3},~\ref{fig:uar-wor-wop4},~\ref{fig:uar-wor-wop5},~\ref{fig:uar-wor-wop6},~\ref{fig:uar-wor-wop7}.\\
Sampling with replacement:  Figure~\ref{fig:uar-wr-wop1},~\ref{fig:uar-wr-wop2},~\ref{fig:uar-wr-wop3},~\ref{fig:uar-wr-wop4},~\ref{fig:uar-wr-wop5},~\ref{fig:uar-wr-wop6}.

\paragraph{Instances Sampled from Mallows $\varphi$-Model, $\varphi = 0.2$\\ and Single-Peaked Instances Sampled Uniformly at Random}
We give results from instances that were sampled according to Mallows $\varphi$-model with $\varphi = 0.2$ and single-peaked instances that were sampled uniformly at random. These show mostly the same behaviour, except for sampling with replacement and applying pruning.

\textbf{Pruning of confidence intervals after every sample.}\\
Sampling without replacement: Figure ~\ref{fig:msp-wor-wp1},~\ref{fig:msp-wor-wp2},~\ref{fig:msp-wor-wp3},~\ref{fig:msp-wor-wp4}.\\
Sampling with replacement: Figure ~\ref{fig:msp-wr-wp1},~\ref{fig:msp-wr-wp2},~\ref{fig:msp-wr-wp3},~\ref{fig:msp-wr-wp4}.

\textbf{No pruning applied.}\\
Sampling without replacement:  Figure ~\ref{fig:msp-wor-wop1},~\ref{fig:msp-wor-wop2},~\ref{fig:msp-wor-wop3},~\ref{fig:msp-wor-wop4},~\ref{fig:msp-wor-wop5}.\\
Sampling with replacement:  Figure ~\ref{fig:msp-wr-wop1},~\ref{fig:msp-wr-wop2},~\ref{fig:msp-wr-wop3},~\ref{fig:msp-wr-wop4},~\ref{fig:msp-wr-wop5}.


\begin{figure*}
    \centering
     \begin{subfigure}[b]{0.49\textwidth}
         \centering
         \includegraphics[trim={0 0 0 1.5cm},clip,width=\textwidth]{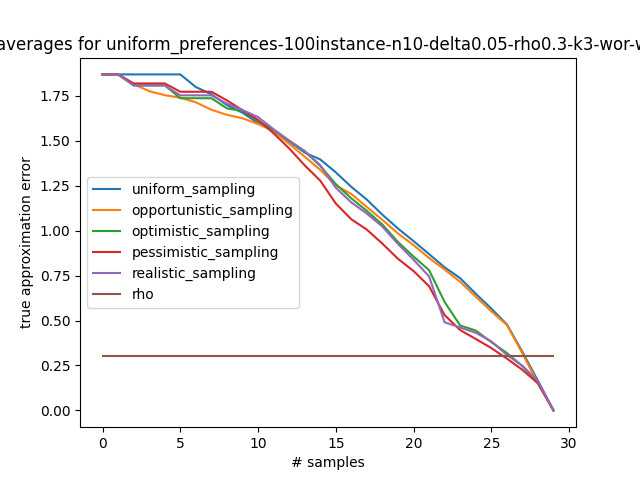}
     \end{subfigure}
     \hfill
     \begin{subfigure}[b]{0.49\textwidth}
         \includegraphics[trim={0 0 0 1.5cm},clip,width=\textwidth]{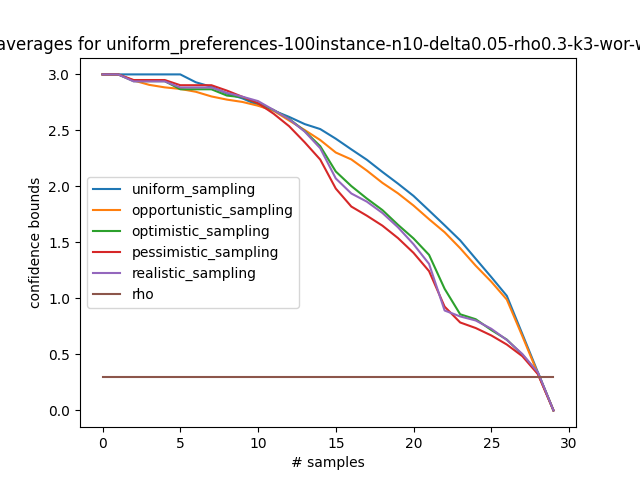}
     \end{subfigure}
    \caption{Sampling Without Replacement: Averages over 100 instances. $\arms=3$, $\rho = 0.3$, $n=10$, $\delta = 0.05$.}
    \label{fig:uar-wor-wp1}
 \end{figure*}
 
\begin{figure*}
    \centering
     \begin{subfigure}[b]{0.49\textwidth}
         \centering
         \includegraphics[trim={0 0 0 1.5cm},clip,width=\textwidth]{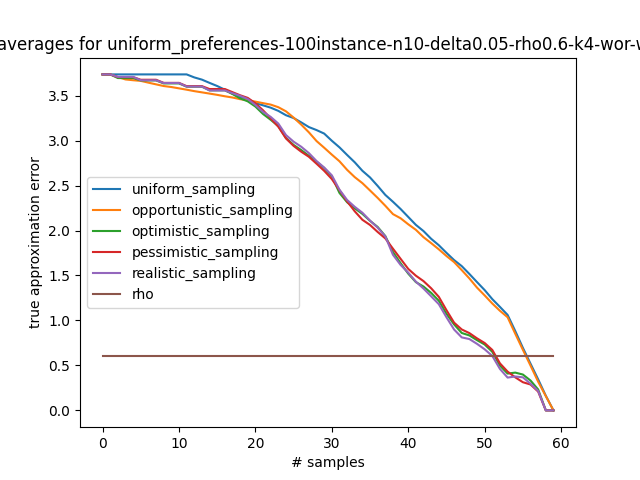}
     \end{subfigure}
     \hfill
     \begin{subfigure}[b]{0.49\textwidth}
         \includegraphics[trim={0 0 0 1.5cm},clip,width=\textwidth]{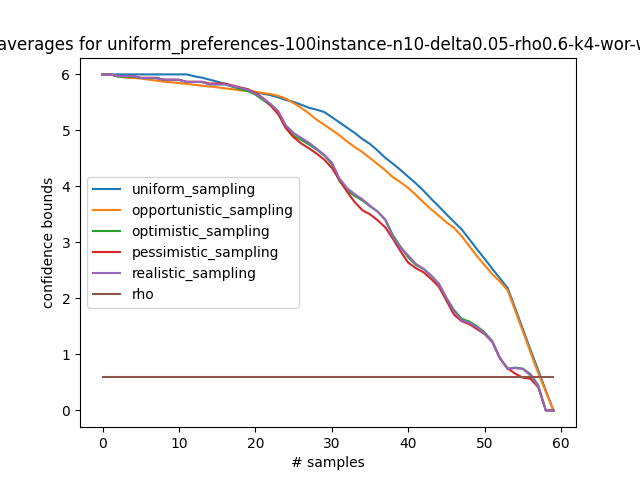}
         
     \end{subfigure}
    \caption{Sampling Without Replacement: Averages over 100 instances. $\arms=4$, $\rho = 0.6$, $n=10$, $\delta = 0.05$.}
    \label{fig:uar-wor-wp2}
 \end{figure*}
 
\begin{figure*}
    \centering
     \begin{subfigure}[b]{0.49\textwidth}
         \centering
         \includegraphics[trim={0 0 0 1.5cm},clip,width=\textwidth]{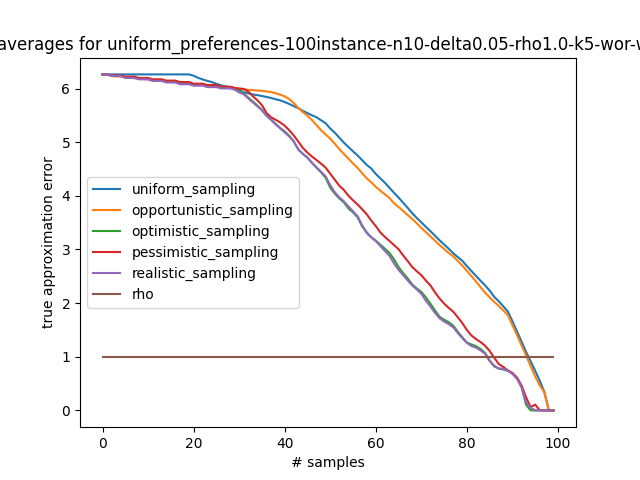}
     \end{subfigure}
     \hfill
     \begin{subfigure}[b]{0.49\textwidth}
         \includegraphics[trim={0 0 0 1.5cm},clip,width=\textwidth]{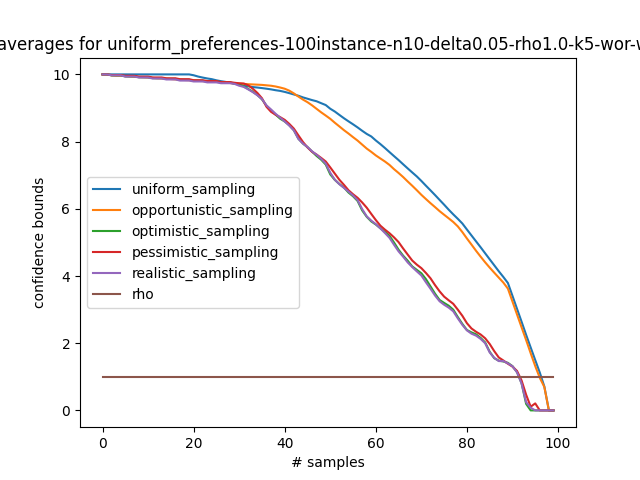}
         
     \end{subfigure}
    \caption{Sampling Without Replacement: Averages over 100 instances. $\arms=5$, $\rho = 1.0$, $n=10$, $\delta = 0.05$.}
    \label{fig:uar-wor-wp3}
 \end{figure*}
 
\begin{figure*}
    \centering
     \begin{subfigure}[b]{0.49\textwidth}
         \centering
         \includegraphics[trim={0 0 0 1.5cm},clip,width=\textwidth]{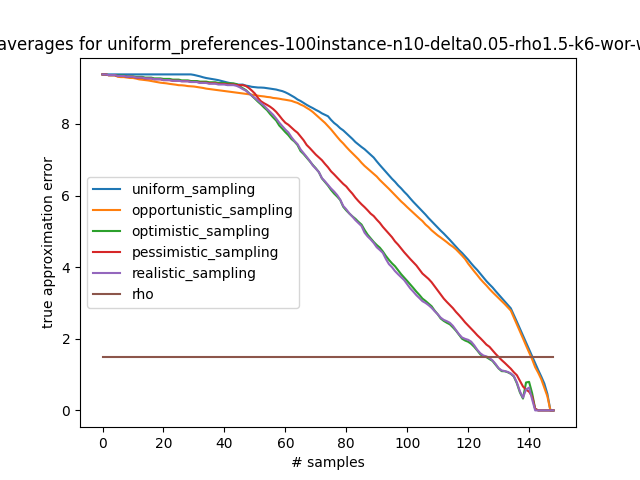}
     \end{subfigure}
     \hfill
     \begin{subfigure}[b]{0.49\textwidth}
         \includegraphics[trim={0 0 0 1.5cm},clip,width=\textwidth]{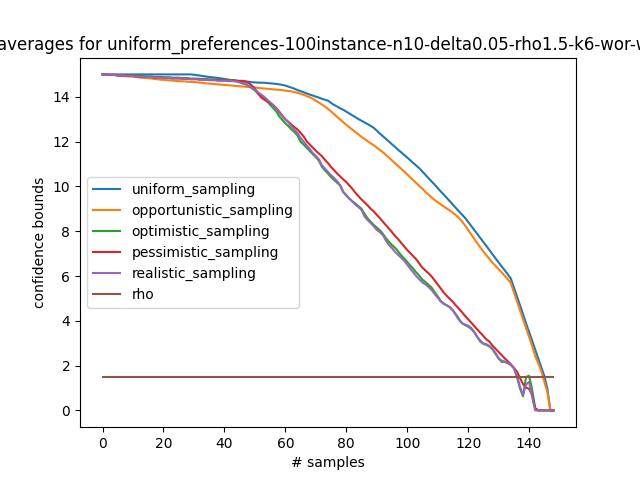}
         
     \end{subfigure}
    \caption{Sampling Without Replacement: Averages over 100 instances. $\arms=6$, $\rho = 1.5$, $n=10$, $\delta = 0.05$.}
    \label{fig:uar-wor-wp4}
 \end{figure*}
 
\begin{figure*}
    \centering
     \begin{subfigure}[b]{0.49\textwidth}
         \centering
         \includegraphics[trim={0 0 0 1.5cm},clip,width=\textwidth]{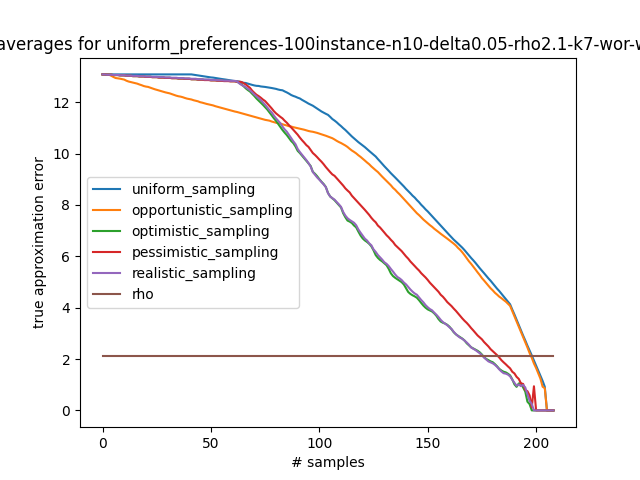}
     \end{subfigure}
     \hfill
     \begin{subfigure}[b]{0.49\textwidth}
         \includegraphics[trim={0 0 0 1.5cm},clip,width=\textwidth]{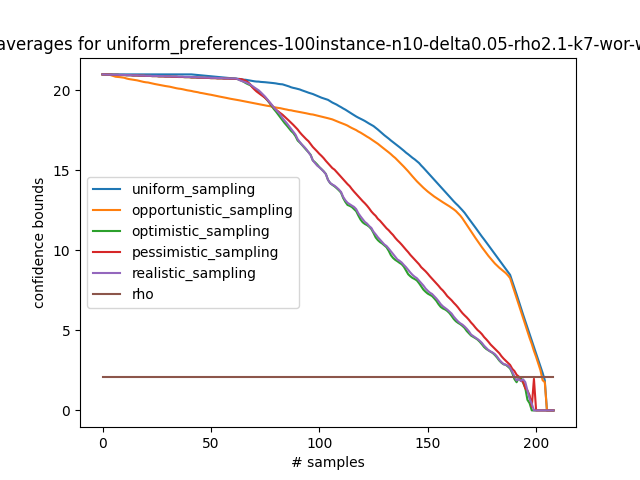}
         
     \end{subfigure}
    \caption{Sampling Without Replacement: Averages over 100 instances. $\arms=7$, $\rho = 2.1$, $n=10$, $\delta = 0.05$.}
    \label{fig:uar-wor-wp5}
 \end{figure*}
 
\begin{figure*}
    \centering
     \begin{subfigure}[b]{0.49\textwidth}
         \centering
         \includegraphics[trim={0 0 0 1.5cm},clip,width=\textwidth]{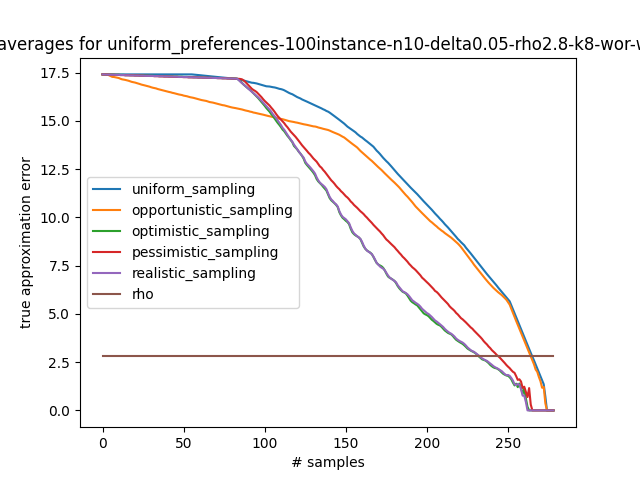}
     \end{subfigure}
     \hfill
     \begin{subfigure}[b]{0.49\textwidth}
         \includegraphics[trim={0 0 0 1.5cm},clip,width=\textwidth]{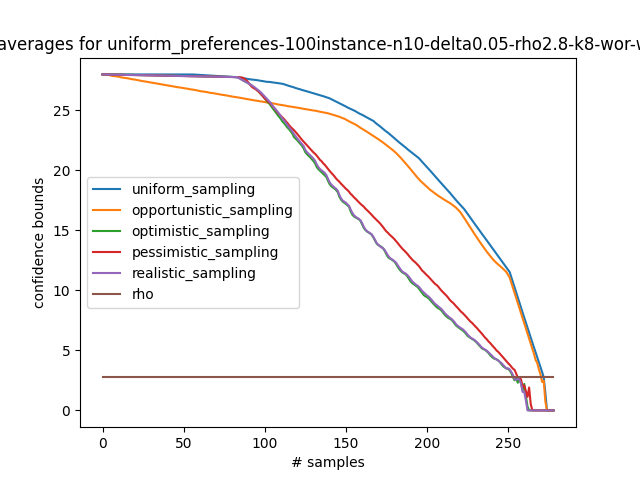}
         
     \end{subfigure}
    \caption{Sampling Without Replacement: Averages over 100 instances. $\arms=8$, $\rho = 2.8$, $n=10$, $\delta = 0.05$.}
    \label{fig:uar-wor-wp6}
 \end{figure*}
 
\begin{figure*}
    \centering
     \begin{subfigure}[b]{0.49\textwidth}
         \centering
         \includegraphics[trim={0 0 0 1.5cm},clip,width=\textwidth]{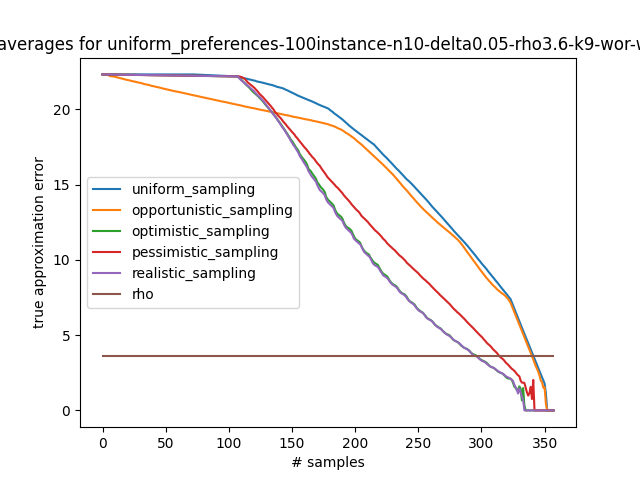}
     \end{subfigure}
     \hfill
     \begin{subfigure}[b]{0.49\textwidth}
         \includegraphics[trim={0 0 0 1.5cm},clip,width=\textwidth]{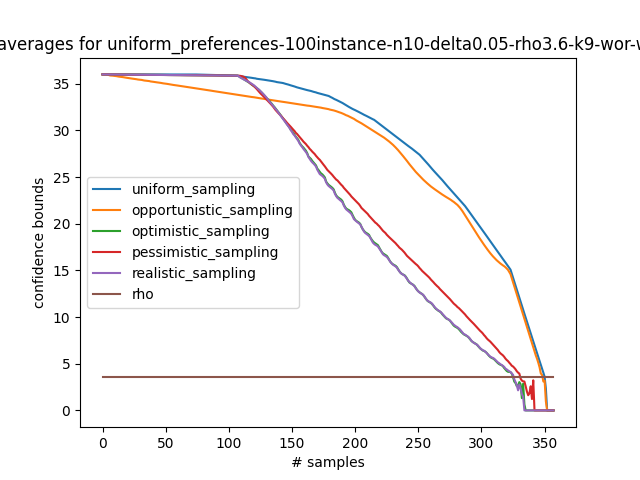}
         
     \end{subfigure}
    \caption{Sampling Without Replacement: Averages over 100 instances. $\arms=9$, $\rho = 3.6$, $n=10$, $\delta = 0.05$.}
    \label{fig:uar-wor-wp7}
 \end{figure*}


\begin{figure*}
    \centering
     \begin{subfigure}[b]{0.49\textwidth}
         \centering
         \includegraphics[trim={0 0 0 1.5cm},clip,width=\textwidth]         {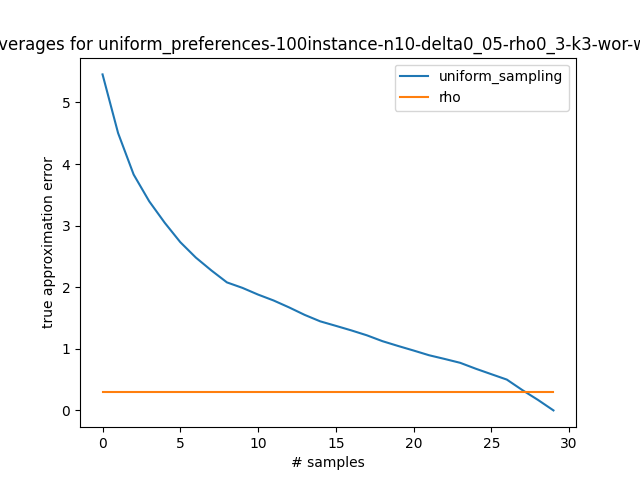}
     \end{subfigure}
     \hfill
     \begin{subfigure}[b]{0.49\textwidth}
         \includegraphics[trim={0 0 0 1.5cm},clip,width=\textwidth]{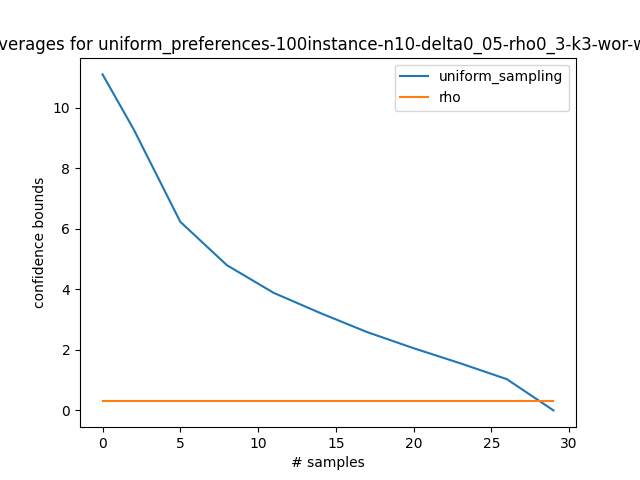}
         
     \end{subfigure}
    \caption{Sampling Without Replacement: Averages over 100 instances. $\arms=3$, $\rho = 0.3$, $n=10$, $\delta = 0.05$.}
    \label{fig:uar-wor-wop1}
 \end{figure*}
 
\begin{figure*}
    \centering
     \begin{subfigure}[b]{0.49\textwidth}
         \centering
         \includegraphics[trim={0 0 0 1.5cm},clip,width=\textwidth]{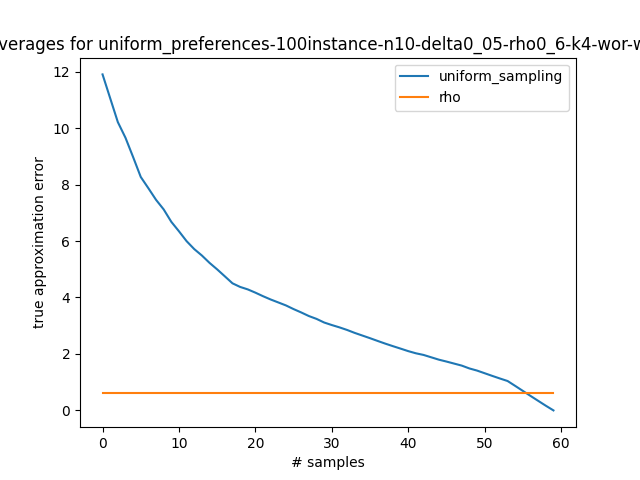}
     \end{subfigure}
     \hfill
     \begin{subfigure}[b]{0.49\textwidth}
         \includegraphics[trim={0 0 0 1.5cm},clip,width=\textwidth]{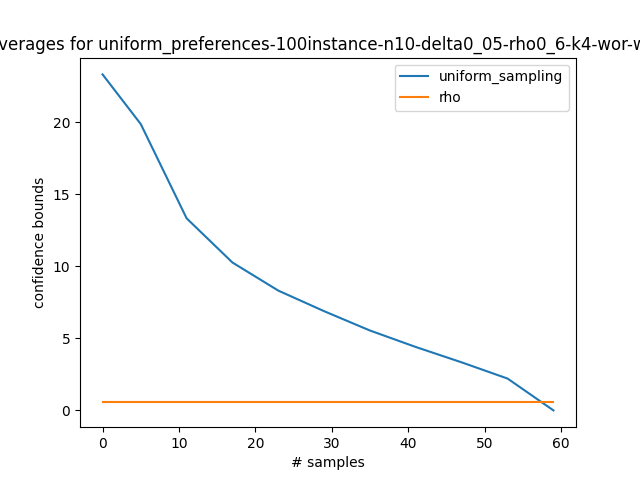}
         
     \end{subfigure}
    \caption{Sampling Without Replacement: Averages over 100 instances. $\arms=4$, $\rho = 0.6$, $n=10$, $\delta = 0.05$.}
    \label{fig:uar-wor-wop2}
 \end{figure*}
 
\begin{figure*}
    \centering
     \begin{subfigure}[b]{0.49\textwidth}
         \centering
         \includegraphics[trim={0 0 0 1.5cm},clip,width=\textwidth]{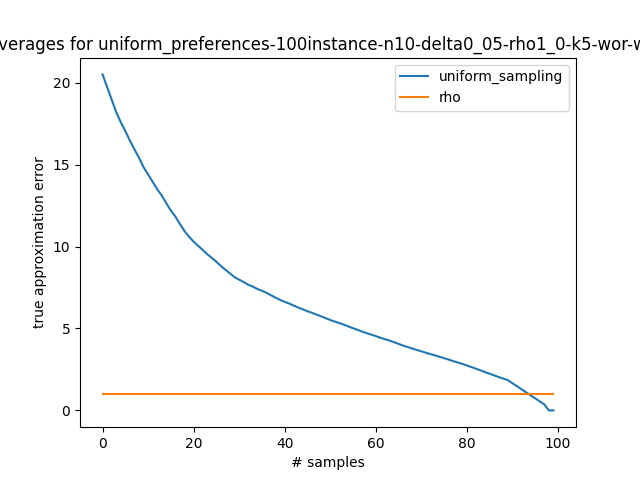}
     \end{subfigure}
     \hfill
     \begin{subfigure}[b]{0.49\textwidth}
         \includegraphics[trim={0 0 0 1.5cm},clip,width=\textwidth]{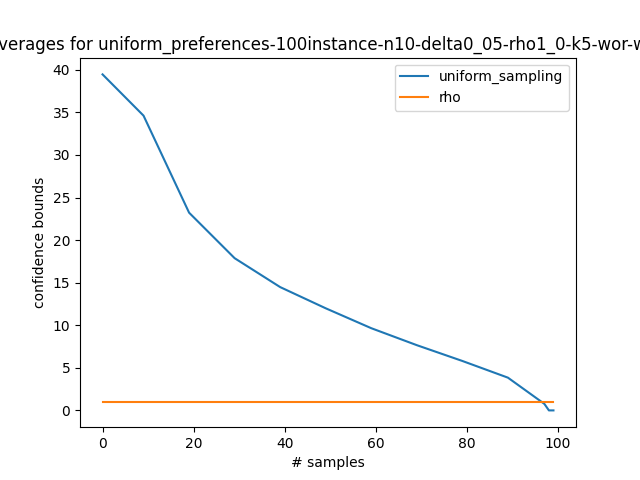}
         
     \end{subfigure}
    \caption{Sampling Without Replacement: Averages over 100 instances. $\arms=5$, $\rho = 1.0$, $n=10$, $\delta = 0.05$.}
    \label{fig:uar-wor-wop3}
 \end{figure*}
 
\begin{figure*}
    \centering
     \begin{subfigure}[b]{0.49\textwidth}
         \centering
         \includegraphics[trim={0 0 0 1.5cm},clip,width=\textwidth]{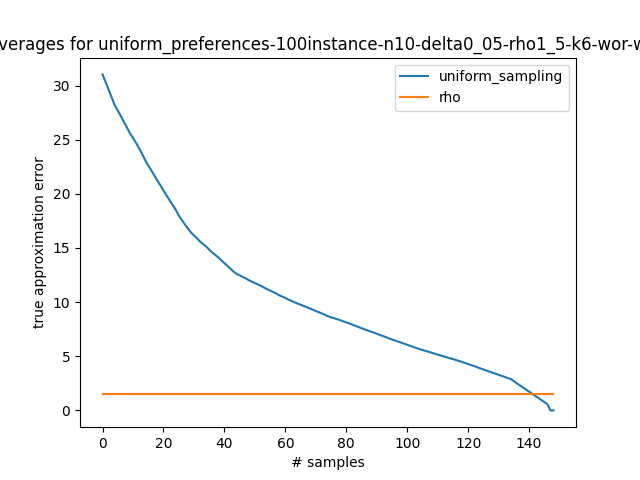}
     \end{subfigure}
     \hfill
     \begin{subfigure}[b]{0.49\textwidth}
         \includegraphics[trim={0 0 0 1.5cm},clip,width=\textwidth]{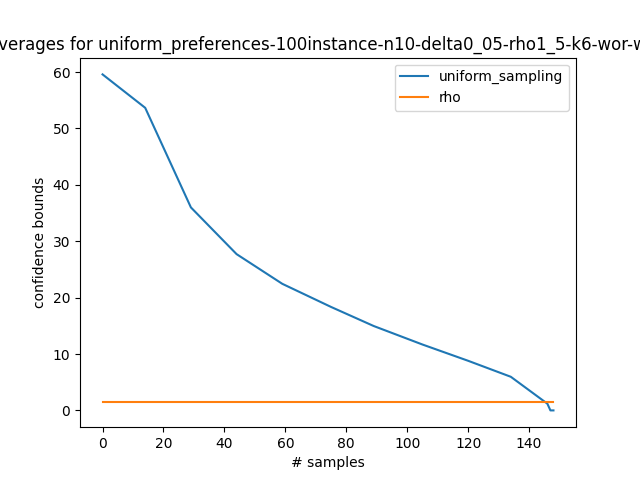}
         
     \end{subfigure}
    \caption{Sampling Without Replacement: Averages over 100 instances. $\arms=6$, $\rho = 1.5$, $n=10$, $\delta = 0.05$.}
    \label{fig:uar-wor-wop4}
 \end{figure*}
 
\begin{figure*}
    \centering
     \begin{subfigure}[b]{0.49\textwidth}
         \centering
         \includegraphics[trim={0 0 0 1.5cm},clip,width=\textwidth]{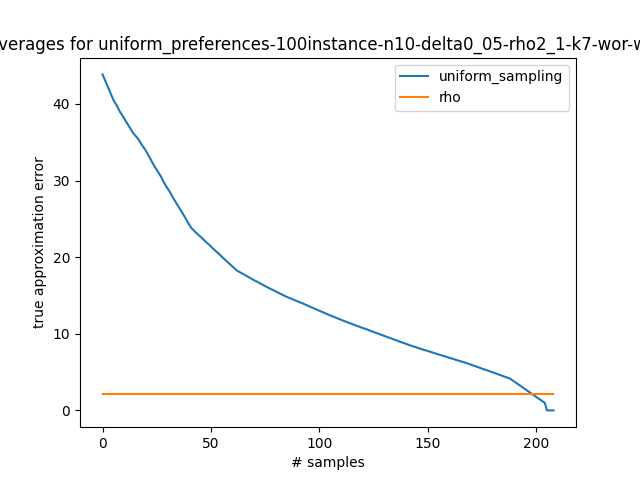}
     \end{subfigure}
     \hfill
     \begin{subfigure}[b]{0.49\textwidth}
         \includegraphics[trim={0 0 0 1.5cm},clip,width=\textwidth]{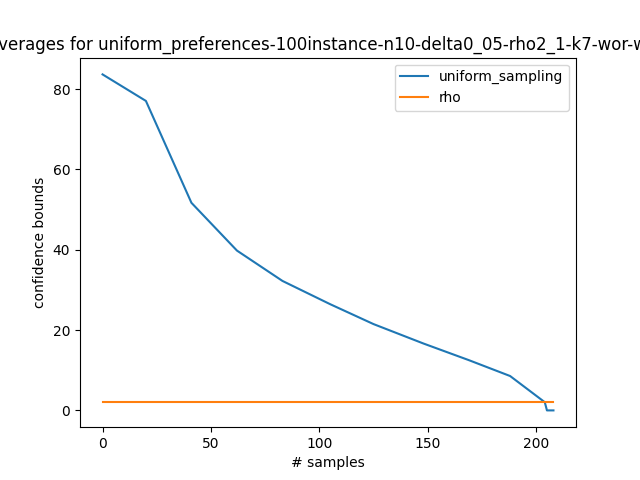}
         
     \end{subfigure}
    \caption{Sampling Without Replacement: Averages over 100 instances. $\arms=7$, $\rho = 2.1$, $n=10$, $\delta = 0.05$.}
    \label{fig:uar-wor-wop5}
 \end{figure*}
 
\begin{figure*}
    \centering
     \begin{subfigure}[b]{0.49\textwidth}
         \centering
         \includegraphics[trim={0 0 0 1.5cm},clip,width=\textwidth]{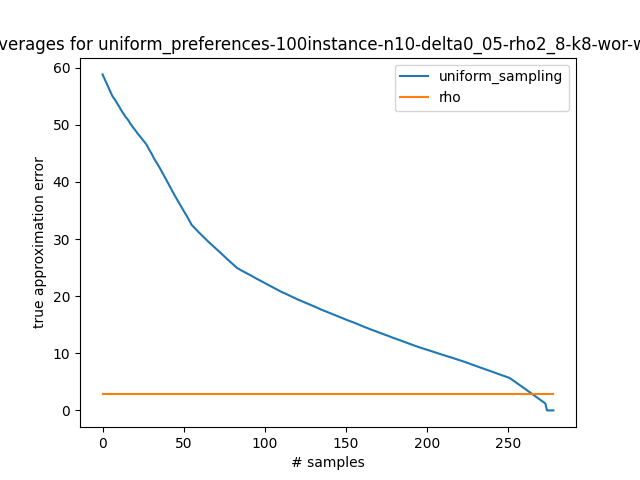}
     \end{subfigure}
     \hfill
     \begin{subfigure}[b]{0.49\textwidth}
         \includegraphics[trim={0 0 0 1.5cm},clip,width=\textwidth]{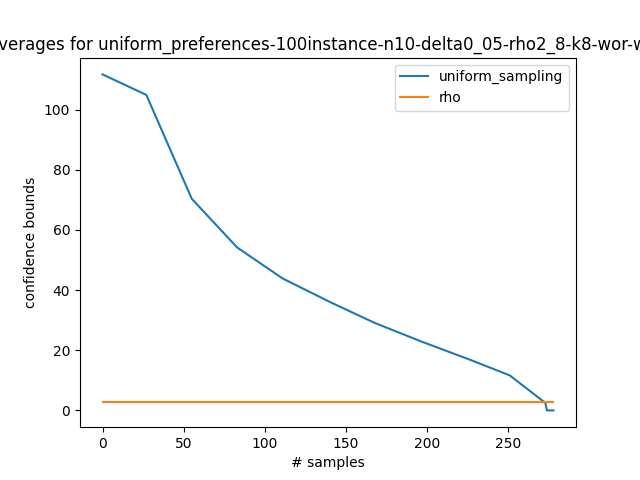}
         
     \end{subfigure}
    \caption{Sampling Without Replacement: Averages over 100 instances. $\arms=8$, $\rho = 2.8$, $n=10$, $\delta = 0.05$.}
    \label{fig:uar-wor-wop6}
 \end{figure*}
 
\begin{figure*}
    \centering
     \begin{subfigure}[b]{0.49\textwidth}
         \centering
         \includegraphics[trim={0 0 0 1.5cm},clip,width=\textwidth]{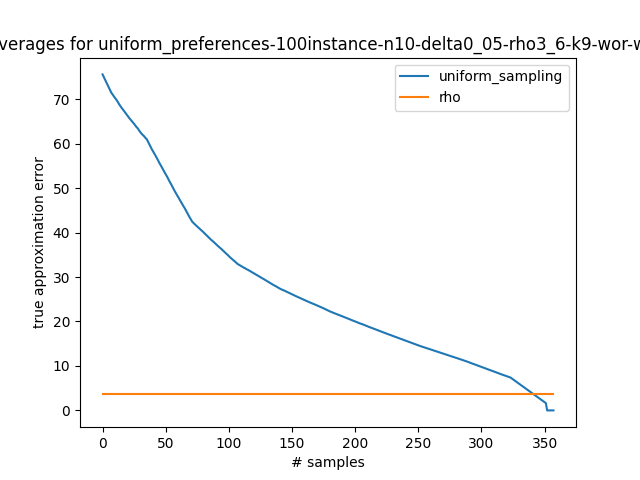}
     \end{subfigure}
     \hfill
     \begin{subfigure}[b]{0.49\textwidth}
         \includegraphics[trim={0 0 0 1.5cm},clip,width=\textwidth]{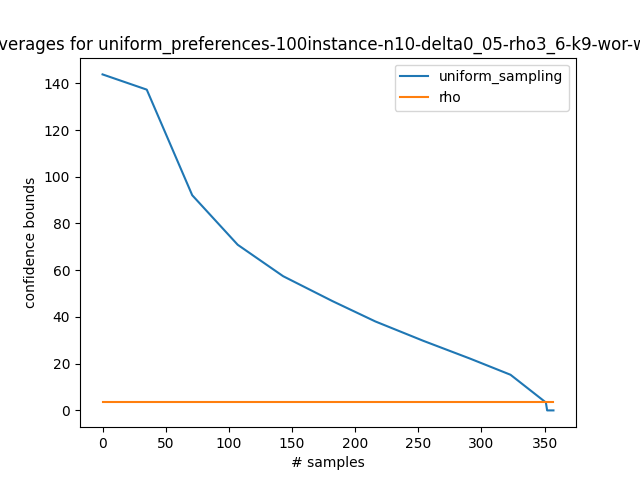}
         
     \end{subfigure}
    \caption{Sampling Without Replacement: Averages over 100 instances. $\arms=9$, $\rho = 3.6$, $n=10$, $\delta = 0.05$.}
    \label{fig:uar-wor-wop7}
 \end{figure*}


\begin{figure*}
    \centering
     \begin{subfigure}[b]{0.49\textwidth}
         \includegraphics[trim={0 0 0 1.5cm},clip,width=\textwidth]{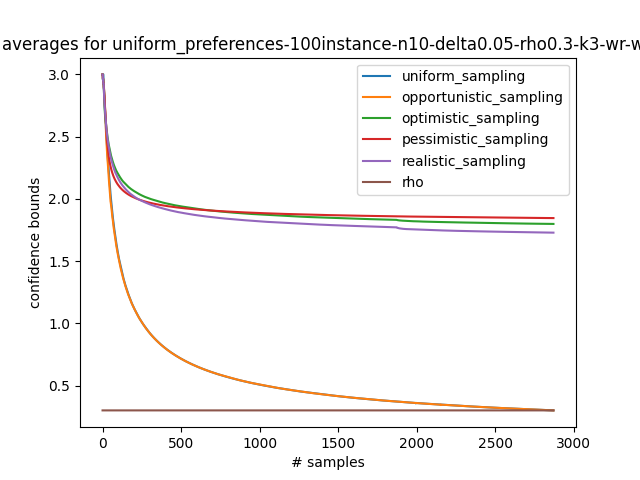}
         
     \end{subfigure}
    \caption{Sampling With Replacement: Averages over 100 instances. $\arms=3$, $\rho = 0.3$, $n=10$, $\delta = 0.05$.}
    \label{fig:uar-wr-wp1}
 \end{figure*}
 
\begin{figure*}
    \centering
     \begin{subfigure}[b]{0.49\textwidth}
         \includegraphics[trim={0 0 0 1.5cm},clip,width=\textwidth]{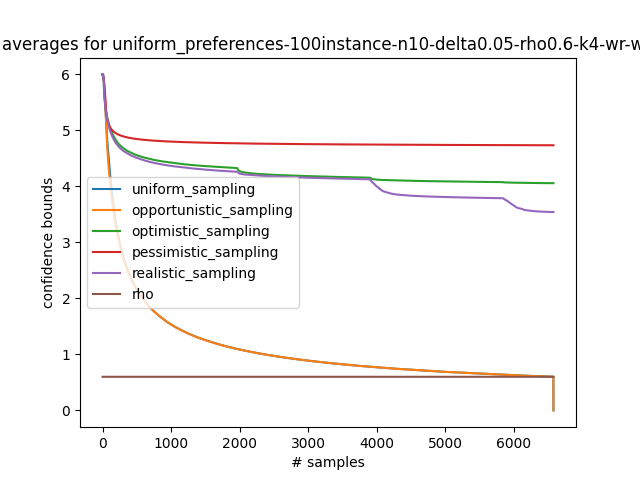}
         
     \end{subfigure}
    \caption{Sampling With Replacement: Averages over 100 instances. $\arms=4$, $\rho = 0.6$, $n=10$, $\delta = 0.05$.}
    \label{fig:uar-wr-wp2}
 \end{figure*}
 
\begin{figure*}
    \centering
     \begin{subfigure}[b]{0.49\textwidth}
         \includegraphics[trim={0 0 0 1.5cm},clip,width=\textwidth]{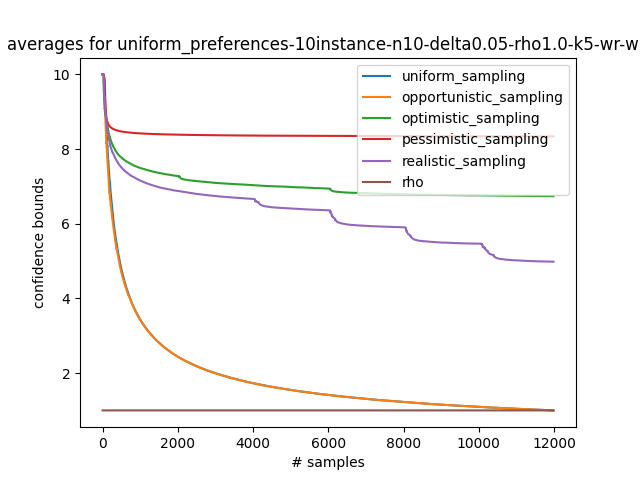}
         
     \end{subfigure}
    \caption{Sampling With Replacement: Averages over 10 instances. $\arms=5$, $\rho = 1.0$, $n=10$, $\delta = 0.05$.}
    \label{fig:uar-wr-wp3}
 \end{figure*}
 
\begin{figure*}
    \centering
     \begin{subfigure}[b]{0.49\textwidth}
         \includegraphics[trim={0 0 0 1.5cm},clip,width=\textwidth]{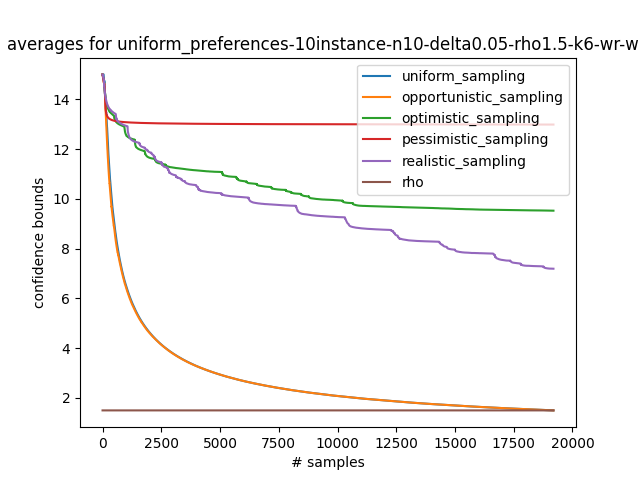}
         
     \end{subfigure}
    \caption{Sampling With Replacement: Averages over 10 instances. $\arms=6$, $\rho = 1.5$, $n=10$, $\delta = 0.05$.}
    \label{fig:uar-wr-wp4}
 \end{figure*}
 
\begin{figure*}
    \centering
     \begin{subfigure}[b]{0.49\textwidth}
         \includegraphics[trim={0 0 0 1.5cm},clip,width=\textwidth]{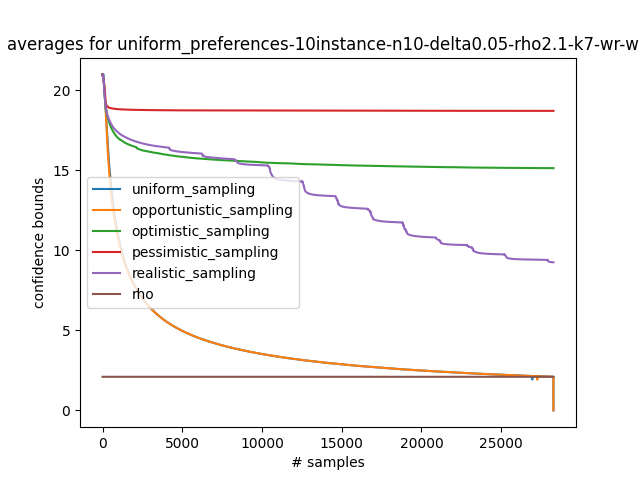}
         
     \end{subfigure}
    \caption{Sampling With Replacement: Averages over 10 instances. $\arms=7$, $\rho = 2.1$, $n=10$, $\delta = 0.05$.}
    \label{fig:uar-wr-wp5}
 \end{figure*}
 

\begin{figure*}
    \centering
     \begin{subfigure}[b]{0.49\textwidth}
         \includegraphics[trim={0 0 0 1.5cm},clip,width=\textwidth]{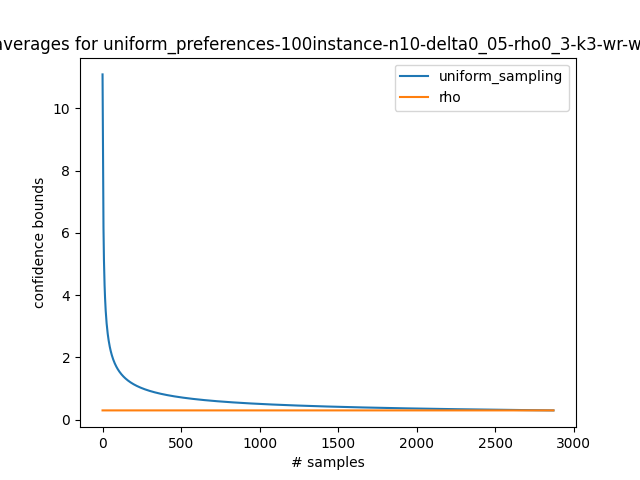}
         
     \end{subfigure}
    \caption{Sampling With Replacement: Averages over 100 instances. $\arms=3$, $\rho = 0.3$, $n=10$, $\delta = 0.05$.}
    \label{fig:uar-wr-wop1}
 \end{figure*}
 
\begin{figure*}
    \centering
     \begin{subfigure}[b]{0.49\textwidth}
         \includegraphics[trim={0 0 0 1.5cm},clip,width=\textwidth]{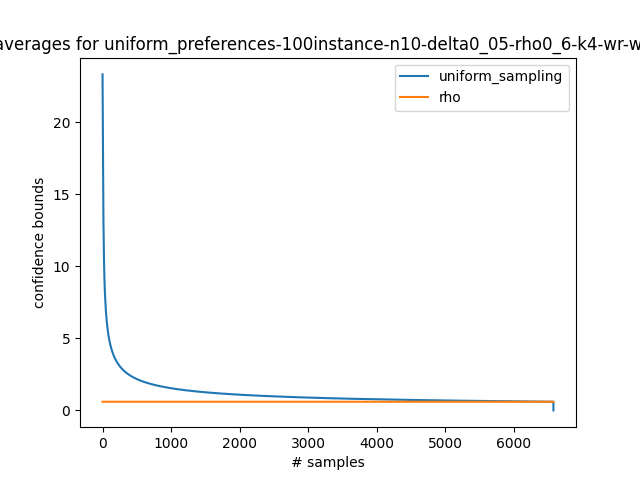}
         
     \end{subfigure}
    \caption{Sampling With Replacement: Averages over 100 instances. $\arms=4$, $\rho = 0.6$, $n=10$, $\delta = 0.05$.}
    \label{fig:uar-wr-wop2}
 \end{figure*}
 
\begin{figure*}
    \centering
     \begin{subfigure}[b]{0.49\textwidth}
         \includegraphics[trim={0 0 0 1.5cm},clip,width=\textwidth]{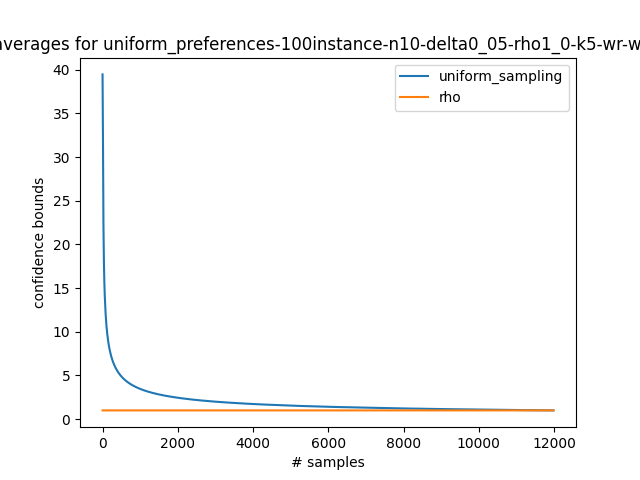}
         
     \end{subfigure}
    \caption{Sampling With Replacement: Averages over 100 instances. $\arms=5$, $\rho = 1.0$, $n=10$, $\delta = 0.05$.}
    \label{fig:uar-wr-wop3}
 \end{figure*}
 
\begin{figure*}
    \centering
     \begin{subfigure}[b]{0.49\textwidth}
         \includegraphics[trim={0 0 0 1.5cm},clip,width=\textwidth]{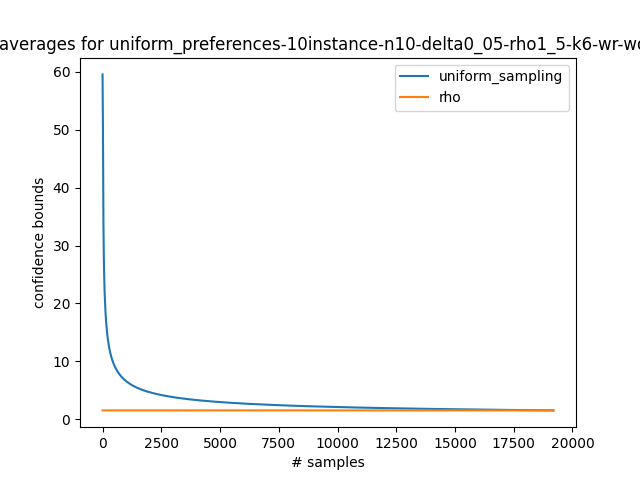}
         
     \end{subfigure}
    \caption{Sampling With Replacement: Averages over 10 instances. $\arms=6$, $\rho = 1.5$, $n=10$, $\delta = 0.05$.}
    \label{fig:uar-wr-wop4}
 \end{figure*}
 
\begin{figure*}
    \centering
     \begin{subfigure}[b]{0.49\textwidth}
         \includegraphics[trim={0 0 0 1.5cm},clip,width=\textwidth]{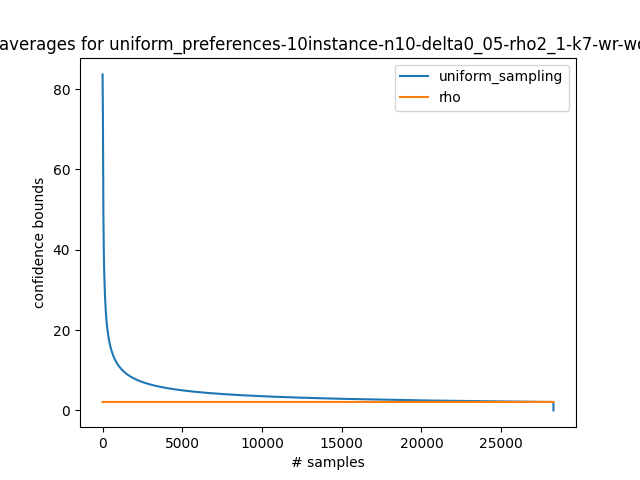}
         
     \end{subfigure}
    \caption{Sampling With Replacement: Averages over 10 instances. $\arms=7$, $\rho = 2.1$, $n=10$, $\delta = 0.05$.}
    \label{fig:uar-wr-wop5}
 \end{figure*}
 
\begin{figure*}
    \centering
     \begin{subfigure}[b]{0.49\textwidth}
         \includegraphics[trim={0 0 0 1.5cm},clip,width=\textwidth]{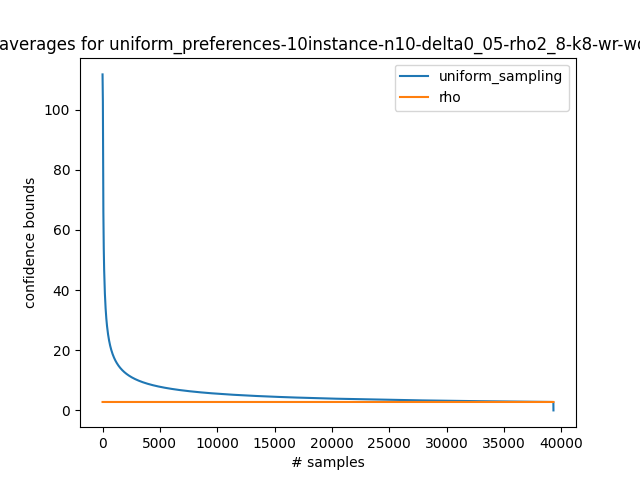}
         
     \end{subfigure}
    \caption{Sampling With Replacement: Averages over 10 instances. $\arms=8$, $\rho = 2.8$, $n=10$, $\delta = 0.05$.}
    \label{fig:uar-wr-wop6}
 \end{figure*}


\begin{figure*}
    \centering
     \begin{subfigure}[b]{0.49\textwidth}
         \centering
         \includegraphics[trim={0 0 0 1.5cm},clip,width=\textwidth]{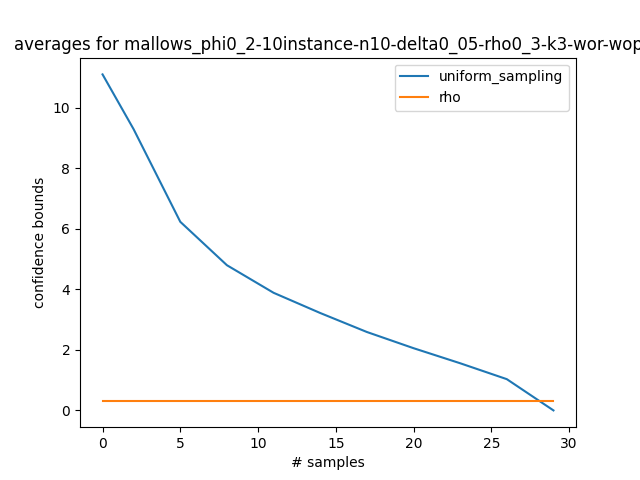}
     \end{subfigure}
     \hfill
     \begin{subfigure}[b]{0.49\textwidth}
         \includegraphics[trim={0 0 0 1.5cm},clip,width=\textwidth]{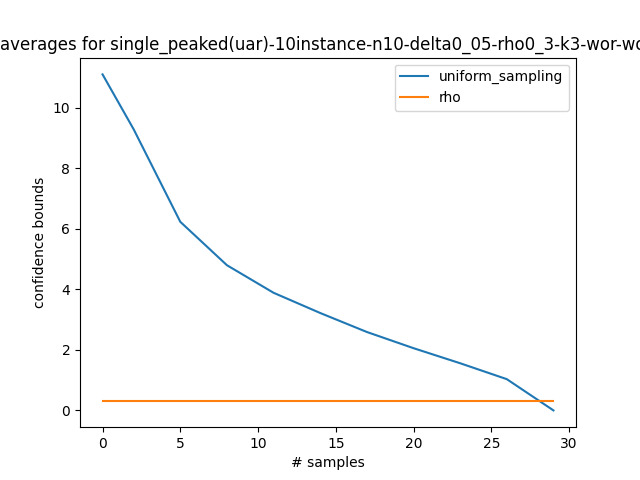}
         
     \end{subfigure}
    \caption{Sampling Without Replacement, without applying pruning: Averages over 10 instances. $\arms=3$, $\rho = 0.3$, $n=10$, $\delta = 0.05$. Left: Instances Sampled from Mallows Model, Right: Single-Peaked Instances.}
    \label{fig:msp-wor-wop1}
 \end{figure*}
 
\begin{figure*}
    \centering
     \begin{subfigure}[b]{0.49\textwidth}
         \centering
         \includegraphics[trim={0 0 0 1.5cm},clip,width=\textwidth]{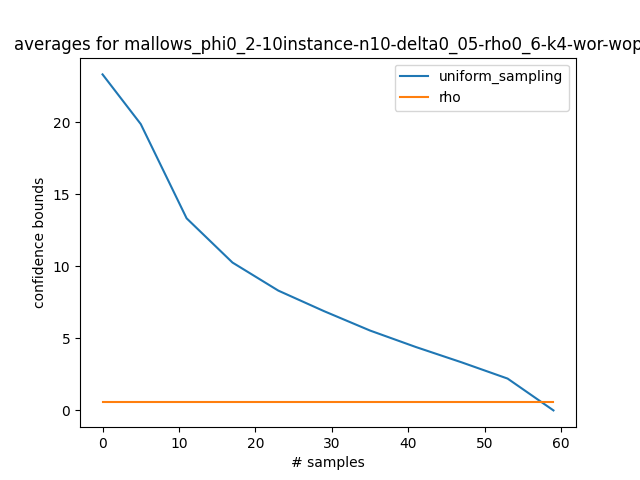}
     \end{subfigure}
     \hfill
     \begin{subfigure}[b]{0.49\textwidth}
         \includegraphics[trim={0 0 0 1.5cm},clip,width=\textwidth]{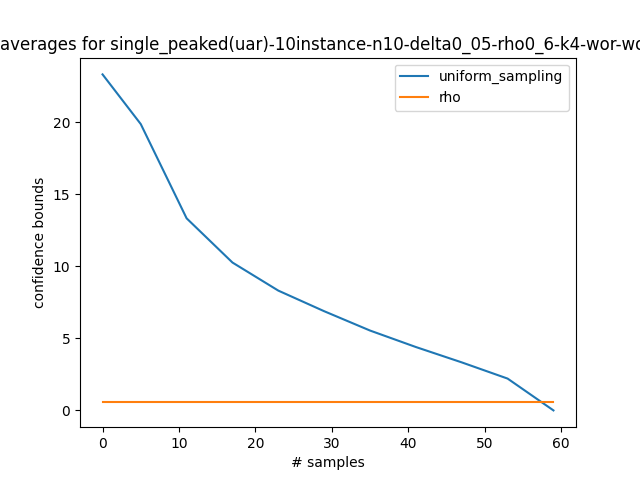}
         
     \end{subfigure}
    \caption{Sampling Without Replacement, without applying pruning: Averages over 10 instances. $\arms=4$, $\rho = 0.6$, $n=10$, $\delta = 0.05$. Left: Instances Sampled from Mallows Model, Right: Single-Peaked Instances.}
    \label{fig:msp-wor-wop2}
 \end{figure*}
 
\begin{figure*}
    \centering
     \begin{subfigure}[b]{0.49\textwidth}
         \centering
         \includegraphics[trim={0 0 0 1.5cm},clip,width=\textwidth]{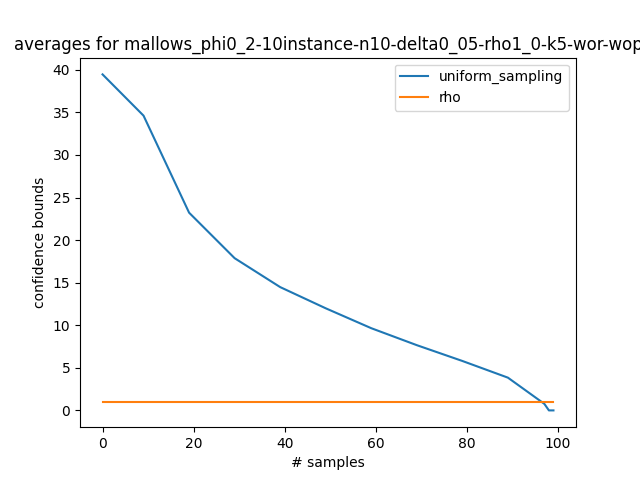}
     \end{subfigure}
     \hfill
     \begin{subfigure}[b]{0.49\textwidth}
         \includegraphics[trim={0 0 0 1.5cm},clip,width=\textwidth]{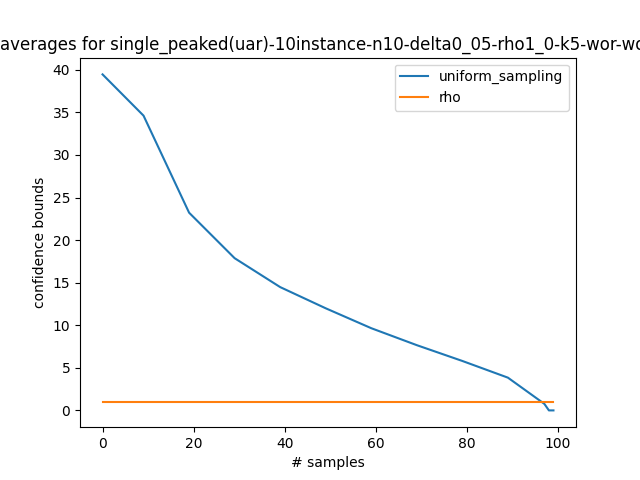}
         
     \end{subfigure}
    \caption{Sampling Without Replacement, without applying pruning: Averages over 10 instances. $\arms=5$, $\rho = 1.0$, $n=10$, $\delta = 0.05$. Left: Instances Sampled from Mallows Model, Right: Single-Peaked Instances.}
    \label{fig:msp-wor-wop3}
 \end{figure*}
 
\begin{figure*}
    \centering
     \begin{subfigure}[b]{0.49\textwidth}
         \centering
         \includegraphics[trim={0 0 0 1.5cm},clip,width=\textwidth]{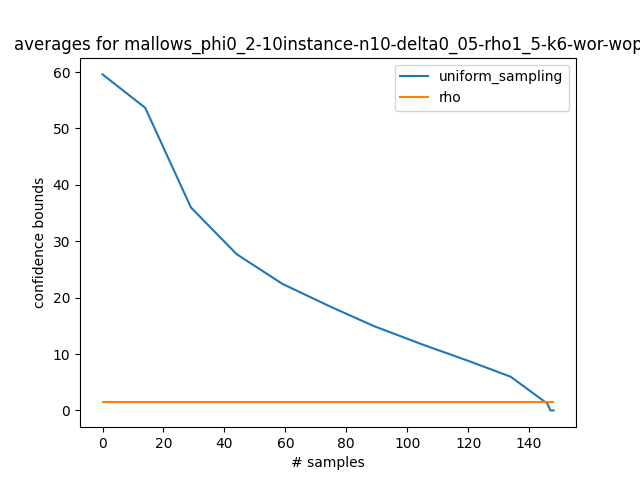}
     \end{subfigure}
     \hfill
     \begin{subfigure}[b]{0.49\textwidth}
         \includegraphics[trim={0 0 0 1.5cm},clip,width=\textwidth]{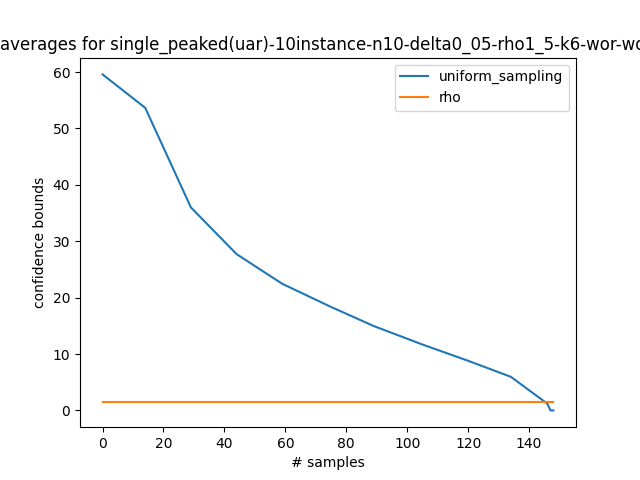}
         
     \end{subfigure}
    \caption{Sampling Without Replacement, without applying pruning: Averages over 10 instances. $\arms=6$, $\rho = 1.5$, $n=10$, $\delta = 0.05$. Left: Instances Sampled from Mallows Model, Right: Single-Peaked Instances.}
    \label{fig:msp-wor-wop4}
 \end{figure*}
 
\begin{figure*}
    \centering
     \begin{subfigure}[b]{0.49\textwidth}
         \centering
         \includegraphics[trim={0 0 0 1.5cm},clip,width=\textwidth]{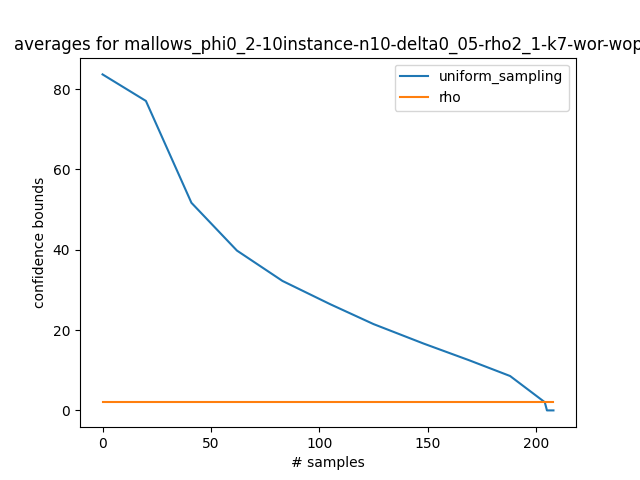}
     \end{subfigure}
     \hfill
     \begin{subfigure}[b]{0.49\textwidth}
         \includegraphics[trim={0 0 0 1.5cm},clip,width=\textwidth]{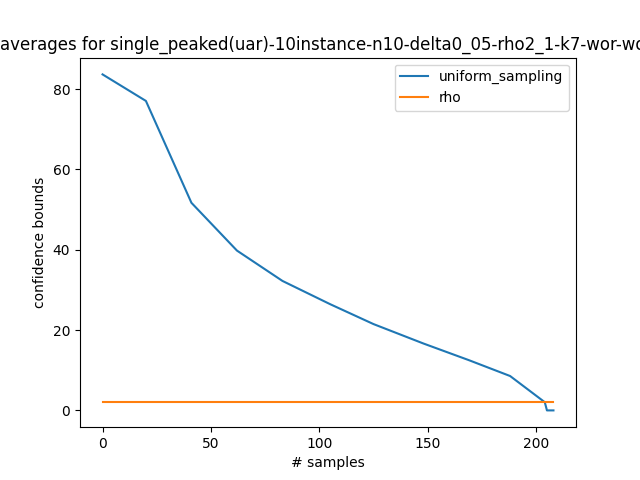}
         
     \end{subfigure}
    \caption{Sampling Without Replacement, without applying pruning: Averages over 10 instances. $\arms=7$, $\rho = 2.1$, $n=10$, $\delta = 0.05$. Left: Instances Sampled from Mallows Model, Right: Single-Peaked Instances.}
    \label{fig:msp-wor-wop5}
 \end{figure*}
 
\begin{figure*}
    \centering
     \begin{subfigure}[b]{0.49\textwidth}
         \centering
         \includegraphics[trim={0 0 0 1.5cm},clip,width=\textwidth]{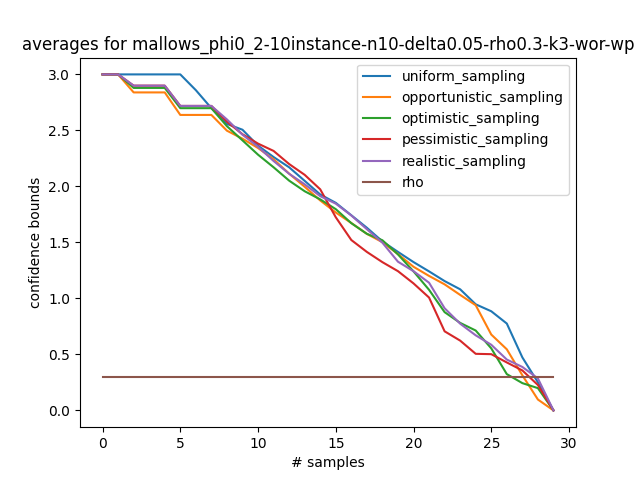}
     \end{subfigure}
     \hfill
     \begin{subfigure}[b]{0.49\textwidth}
         \includegraphics[trim={0 0 0 1.5cm},clip,width=\textwidth]{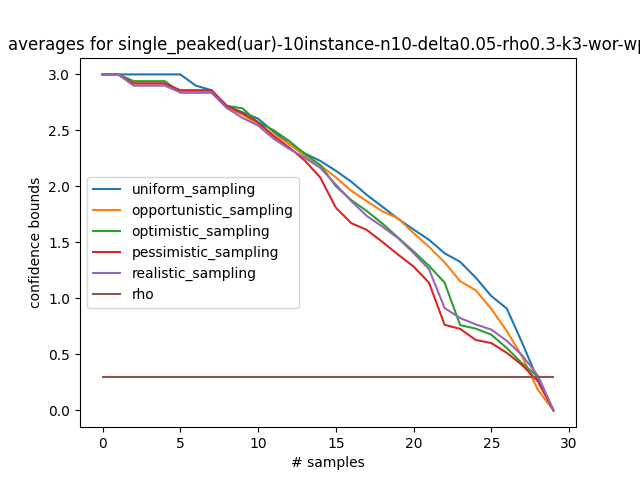}
         
     \end{subfigure}
    \caption{Sampling Without Replacement, pruning applied in every step: Averages over 10 instances. $\arms=3$, $\rho = 0.3$, $n=10$, $\delta = 0.05$. Left: Instances Sampled from Mallows Model, Right: Single-Peaked Instances.}
    \label{fig:msp-wor-wp1}
 \end{figure*}

\begin{figure*}
    \centering
     \begin{subfigure}[b]{0.49\textwidth}
         \centering
         \includegraphics[trim={0 0 0 1.5cm},clip,width=\textwidth]{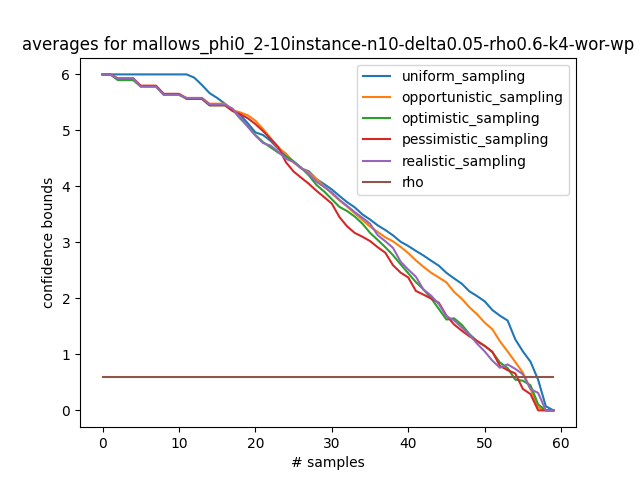}
     \end{subfigure}
     \hfill
     \begin{subfigure}[b]{0.49\textwidth}
         \includegraphics[trim={0 0 0 1.5cm},clip,width=\textwidth]{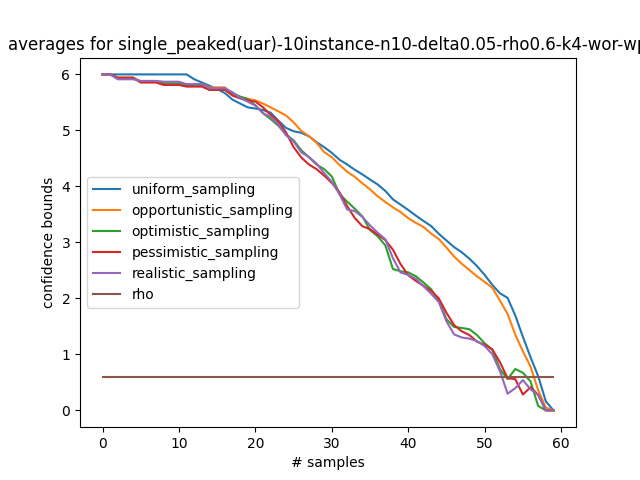}
         
     \end{subfigure}
    \caption{Sampling Without Replacement, pruning applied in every step: Averages over 10 instances. $\arms=4$, $\rho = 0.6$, $n=10$, $\delta = 0.05$. Left: Instances Sampled from Mallows Model, Right: Single-Peaked Instances.}
    \label{fig:msp-wor-wp2}
 \end{figure*}
 
\begin{figure*}
    \centering
     \begin{subfigure}[b]{0.49\textwidth}
         \centering
         \includegraphics[trim={0 0 0 1.5cm},clip,width=\textwidth]{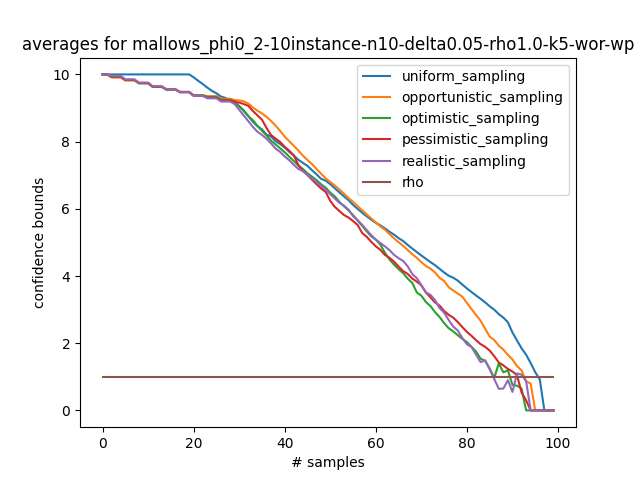}
     \end{subfigure}
     \hfill
     \begin{subfigure}[b]{0.49\textwidth}
         \includegraphics[trim={0 0 0 1.5cm},clip,width=\textwidth]{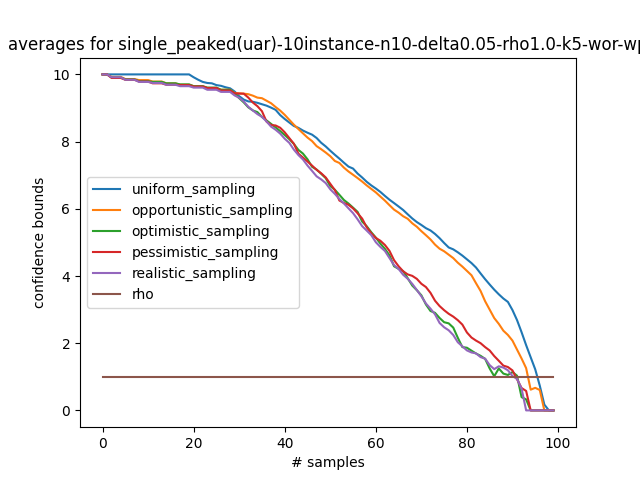}
         
     \end{subfigure}
    \caption{Sampling Without Replacement, pruning applied in every step: Averages over 10 instances. $\arms=5$, $\rho = 1.0$, $n=10$, $\delta = 0.05$. Left: Instances Sampled from Mallows Model, Right: Single-Peaked Instances.}
    \label{fig:msp-wor-wp3}
 \end{figure*}
 
\begin{figure*}
    \centering
     \begin{subfigure}[b]{0.49\textwidth}
         \centering
         \includegraphics[trim={0 0 0 1.5cm},clip,width=\textwidth]{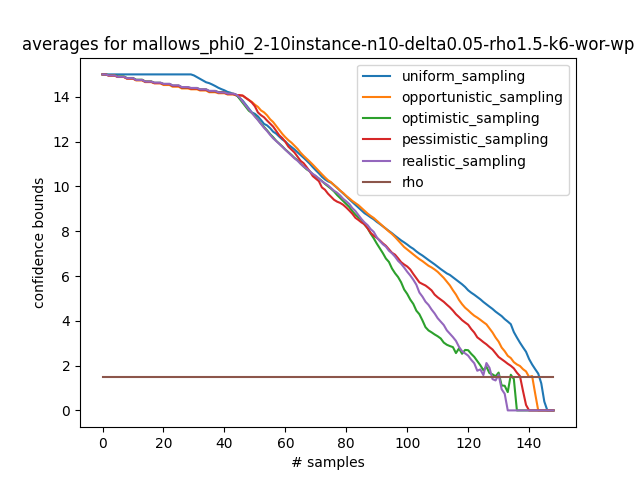}
     \end{subfigure}
     \hfill
     \begin{subfigure}[b]{0.49\textwidth}
         \includegraphics[trim={0 0 0 1.5cm},clip,width=\textwidth]{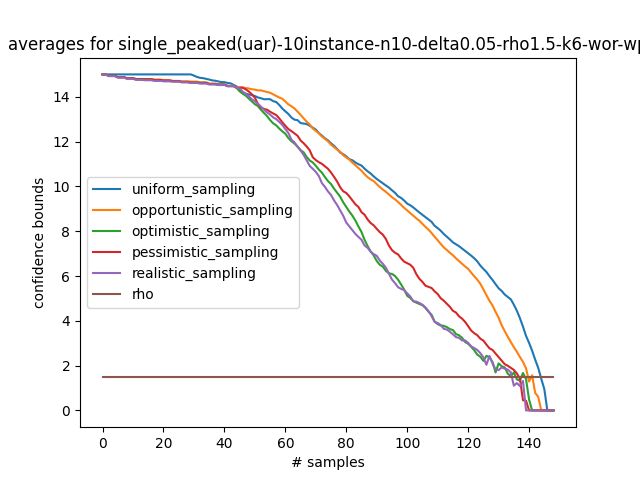}
         
     \end{subfigure}
    \caption{Sampling Without Replacement, pruning applied in every step: Averages over 10 instances. $\arms=6$, $\rho = 1.5$, $n=10$, $\delta = 0.05$. Left: Instances Sampled from Mallows Model, Right: Single-Peaked Instances.}
    \label{fig:msp-wor-wp4}
 \end{figure*}


\begin{figure*}
    \centering
     \begin{subfigure}[b]{0.49\textwidth}
         \centering
         \includegraphics[trim={0 0 0 1.5cm},clip,width=\textwidth]{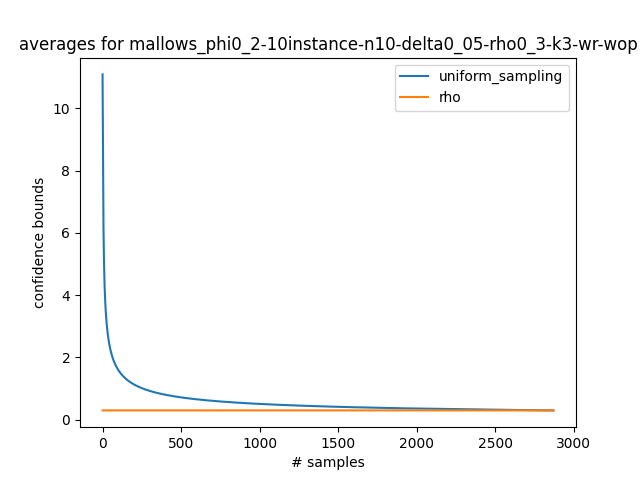}
     \end{subfigure}
     \hfill
     \begin{subfigure}[b]{0.49\textwidth}
         \includegraphics[trim={0 0 0 1.5cm},clip,width=\textwidth]{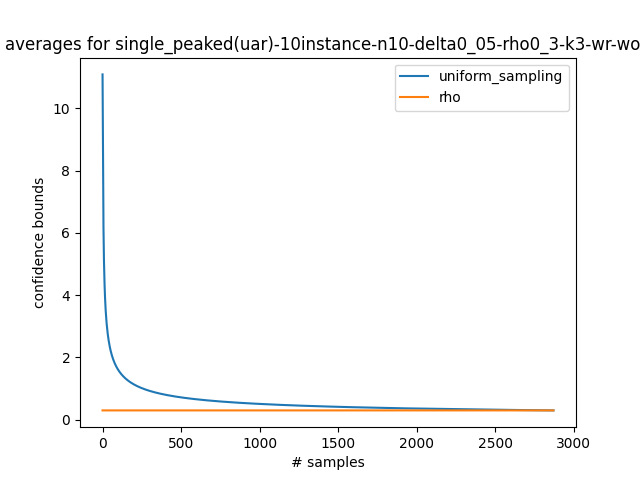}
         
     \end{subfigure}
    \caption{Sampling With Replacement, without applying pruning: Averages over 10 instances. $\arms=3$, $\rho = 0.3$, $n=10$, $\delta = 0.05$. Left: Instances Sampled from Mallows Model, Right: Single-Peaked Instances.}
    \label{fig:msp-wr-wop1}
 \end{figure*}
 
\begin{figure*}
    \centering
     \begin{subfigure}[b]{0.49\textwidth}
         \centering
         \includegraphics[trim={0 0 0 1.5cm},clip,width=\textwidth]{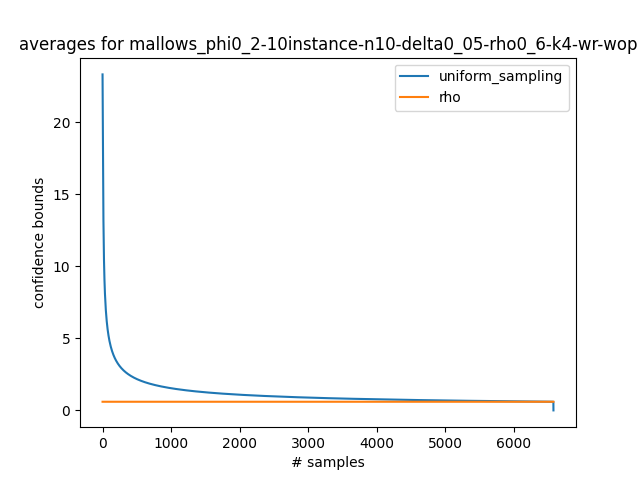}
     \end{subfigure}
     \hfill
     \begin{subfigure}[b]{0.49\textwidth}
         \includegraphics[trim={0 0 0 1.5cm},clip,width=\textwidth]{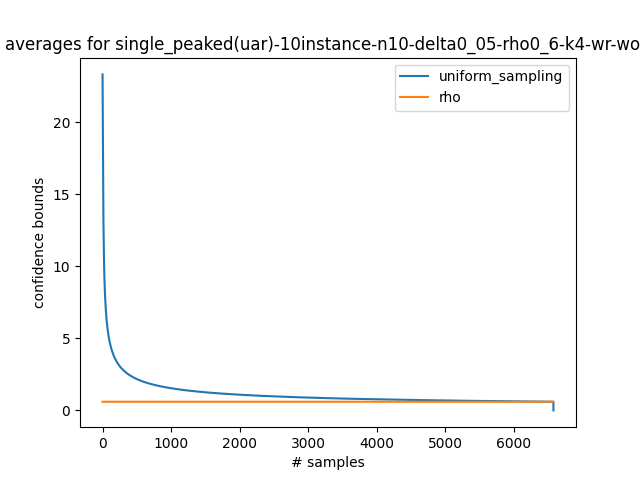}
         
     \end{subfigure}
    \caption{Sampling With Replacement, without applying pruning: Averages over 10 instances. $\arms=4$, $\rho = 0.6$, $n=10$, $\delta = 0.05$. Left: Instances Sampled from Mallows Model, Right: Single-Peaked Instances.}
    \label{fig:msp-wr-wop2}
 \end{figure*}
 
\begin{figure*}
    \centering
     \begin{subfigure}[b]{0.49\textwidth}
         \centering
         \includegraphics[trim={0 0 0 1.5cm},clip,width=\textwidth]{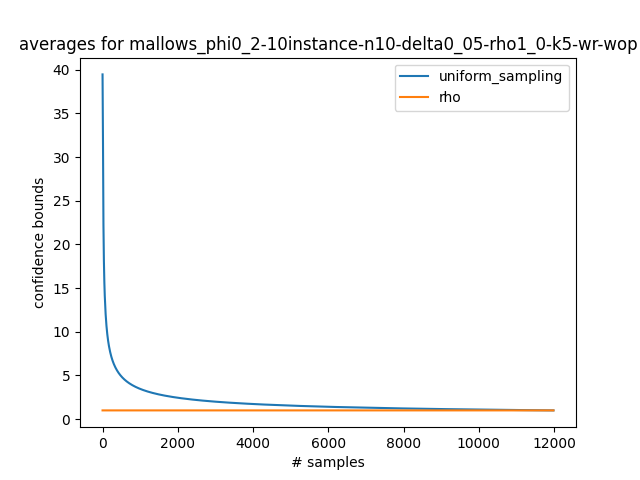}
     \end{subfigure}
     \hfill
     \begin{subfigure}[b]{0.49\textwidth}
         \includegraphics[trim={0 0 0 1.5cm},clip,width=\textwidth]{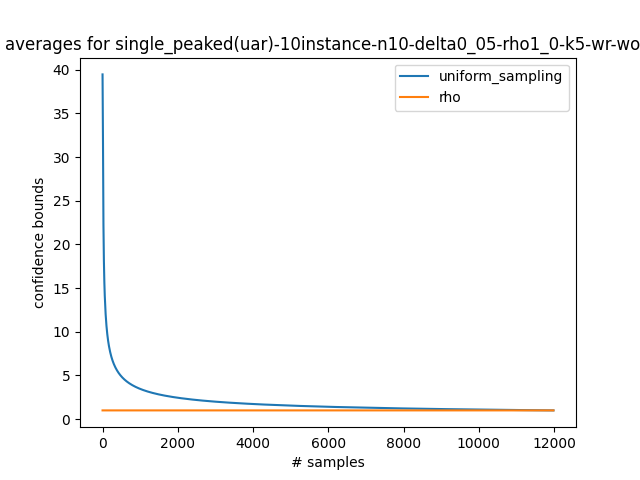}
         
     \end{subfigure}
    \caption{Sampling With Replacement, without applying pruning: Averages over 10 instances. $\arms=5$, $\rho = 1.0$, $n=10$, $\delta = 0.05$. Left: Instances Sampled from Mallows Model, Right: Single-Peaked Instances.}
    \label{fig:msp-wr-wop3}
 \end{figure*}
 
\begin{figure*}
    \centering
     \begin{subfigure}[b]{0.49\textwidth}
         \centering
         \includegraphics[trim={0 0 0 1.5cm},clip,width=\textwidth]{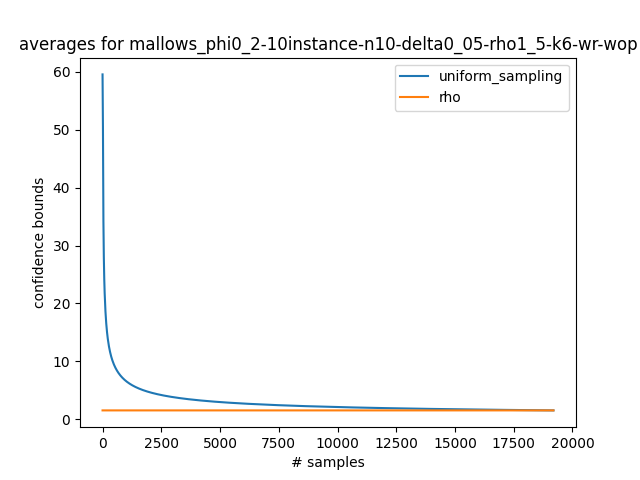}
     \end{subfigure}
     \hfill
     \begin{subfigure}[b]{0.49\textwidth}
         \includegraphics[trim={0 0 0 1.5cm},clip,width=\textwidth]{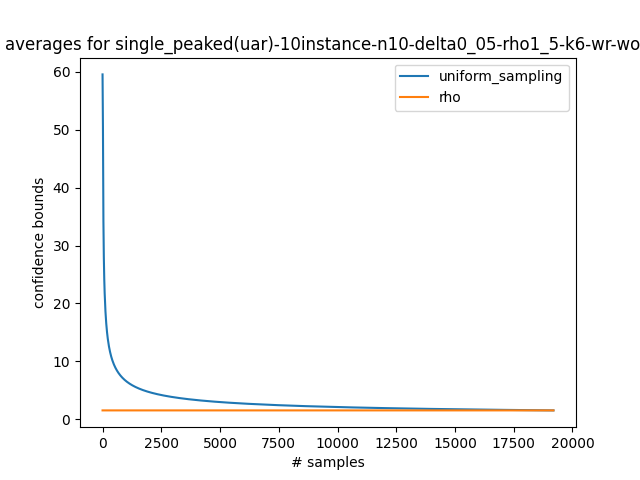}
         
     \end{subfigure}
    \caption{Sampling With Replacement, without applying pruning: Averages over 10 instances. $\arms=6$, $\rho = 1.5$, $n=10$, $\delta = 0.05$. Left: Instances Sampled from Mallows Model, Right: Single-Peaked Instances.}
    \label{fig:msp-wr-wop4}
 \end{figure*}
 
\begin{figure*}
    \centering
     \begin{subfigure}[b]{0.49\textwidth}
         \centering
         \includegraphics[trim={0 0 0 1.5cm},clip,width=\textwidth]{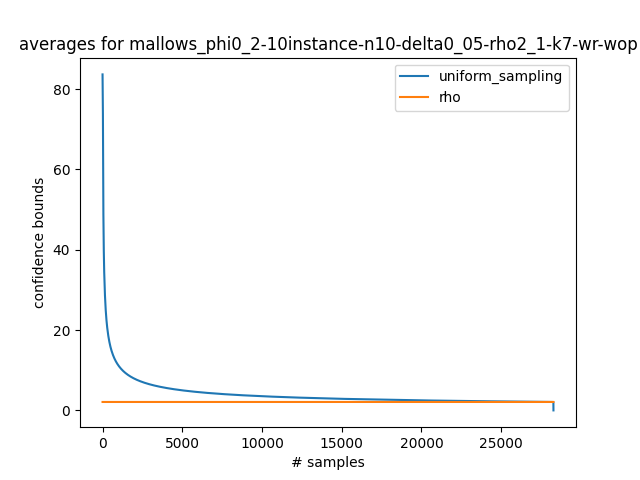}
     \end{subfigure}
     \hfill
     \begin{subfigure}[b]{0.49\textwidth}
         \includegraphics[trim={0 0 0 1.5cm},clip,width=\textwidth]{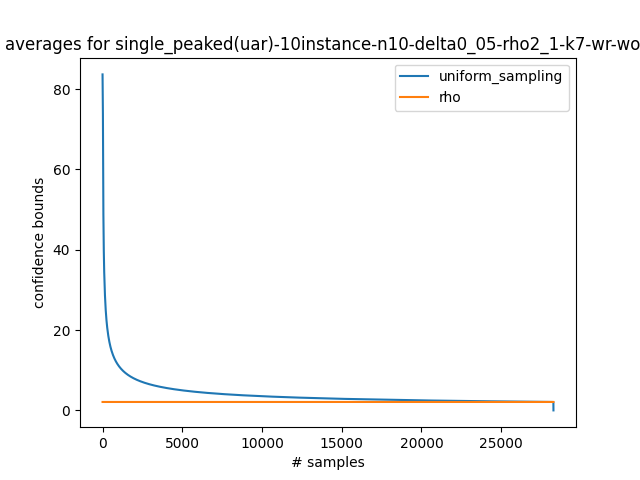}
         
     \end{subfigure}
    \caption{Sampling With Replacement, without applying pruning: Averages over 10 instances. $\arms=7$, $\rho = 2.1$, $n=10$, $\delta = 0.05$. Left: Instances Sampled from Mallows Model, Right: Single-Peaked Instances.}
    \label{fig:msp-wr-wop5}
 \end{figure*}
 
\begin{figure*}
    \centering
     \begin{subfigure}[b]{0.49\textwidth}
         \centering
         \includegraphics[trim={0 0 0 1.5cm},clip,width=\textwidth]{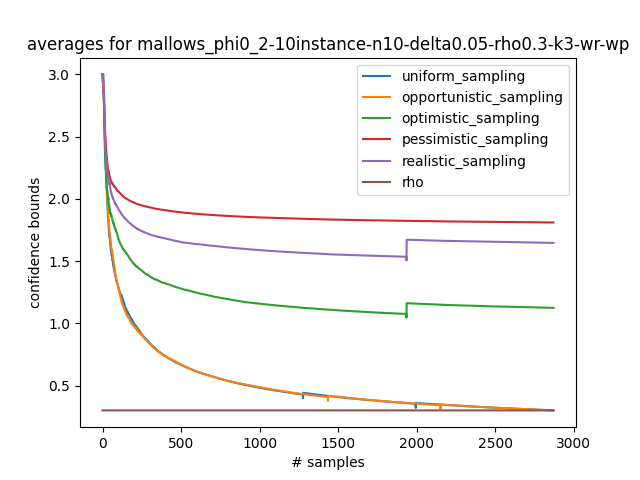}
     \end{subfigure}
     \hfill
     \begin{subfigure}[b]{0.49\textwidth}
         \includegraphics[trim={0 0 0 1.5cm},clip,width=\textwidth]{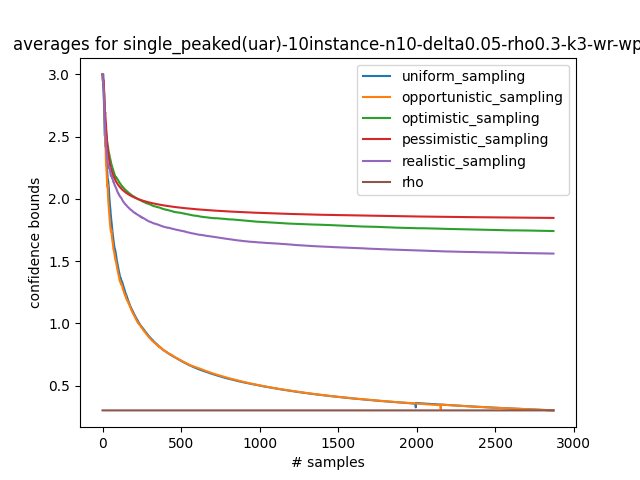}
         
     \end{subfigure}
    \caption{Sampling With Replacement, pruning applied in every step: Averages over 10 instances. $\arms=3$, $\rho = 0.3$, $n=10$, $\delta = 0.05$. Left: Instances Sampled from Mallows Model, Right: Single-Peaked Instances.}
    \label{fig:msp-wr-wp1}
 \end{figure*}

\begin{figure*}
    \centering
     \begin{subfigure}[b]{0.49\textwidth}
         \centering
         \includegraphics[trim={0 0 0 1.5cm},clip,width=\textwidth]{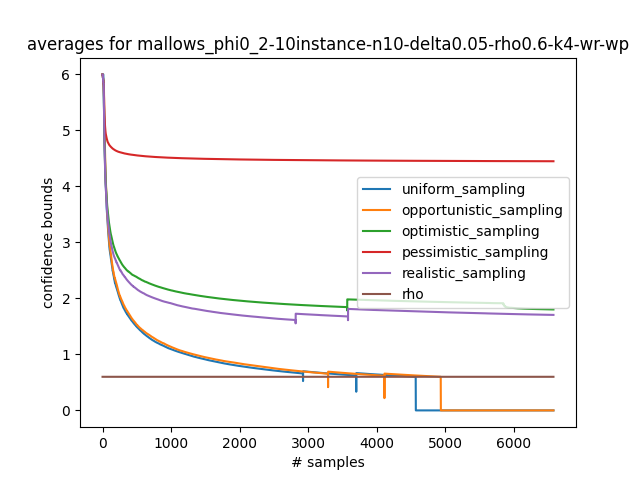}
     \end{subfigure}
     \hfill
     \begin{subfigure}[b]{0.49\textwidth}
         \includegraphics[trim={0 0 0 1.5cm},clip,width=\textwidth]{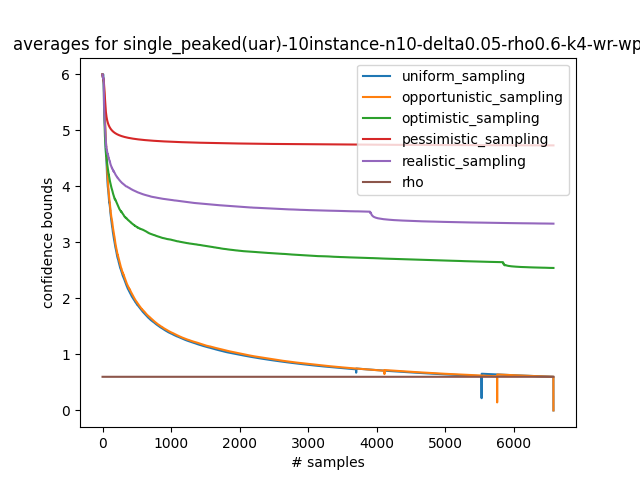}
         
     \end{subfigure}
    \caption{Sampling With Replacement, pruning applied in every step: Averages over 10 instances. $\arms=4$, $\rho = 0.6$, $n=10$, $\delta = 0.05$. Left: Instances Sampled from Mallows Model, Right: Single-Peaked Instances.}
    \label{fig:msp-wr-wp2}
 \end{figure*}
 
\begin{figure*}
    \centering
     \begin{subfigure}[b]{0.49\textwidth}
         \centering
         \includegraphics[trim={0 0 0 1.5cm},clip,width=\textwidth]{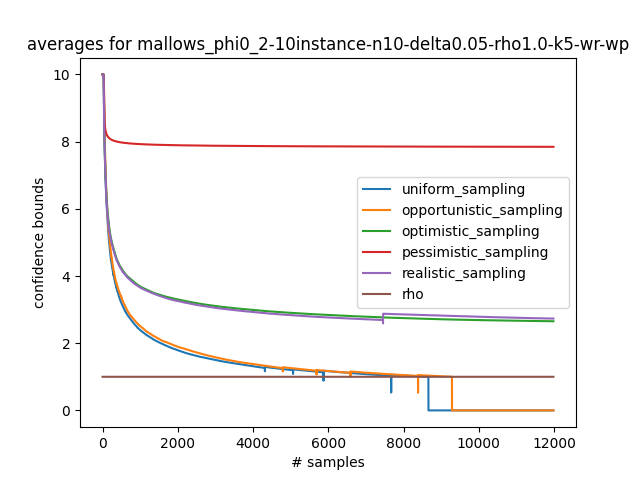}
     \end{subfigure}
     \hfill
     \begin{subfigure}[b]{0.49\textwidth}
         \includegraphics[trim={0 0 0 1.5cm},clip,width=\textwidth]{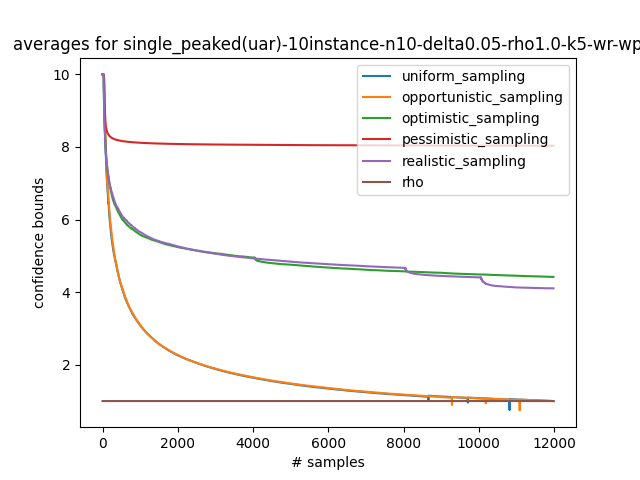}
         
     \end{subfigure}
    \caption{Sampling With Replacement, pruning applied in every step: Averages over 10 instances. $\arms=5$, $\rho = 1.0$, $n=10$, $\delta = 0.05$. Left: Instances Sampled from Mallows Model, Right: Single-Peaked Instances.}
    \label{fig:msp-wr-wp3}
 \end{figure*}
 
\begin{figure*}
    \centering
     \begin{subfigure}[b]{0.49\textwidth}
         \centering
         \includegraphics[trim={0 0 0 1.5cm},clip,width=\textwidth]{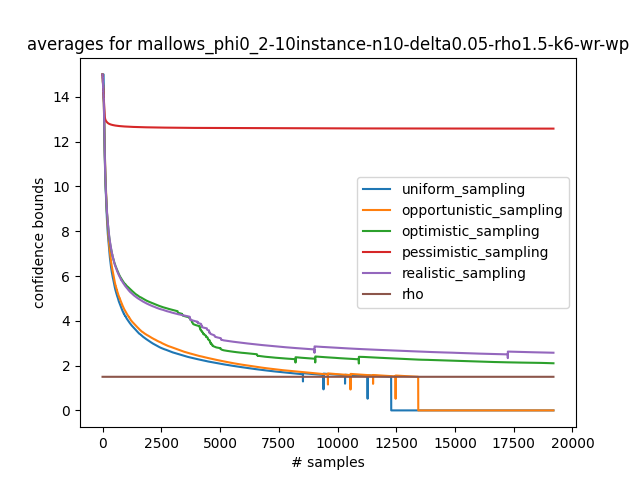}
     \end{subfigure}
     \hfill
     \begin{subfigure}[b]{0.49\textwidth}
         \includegraphics[trim={0 0 0 1.5cm},clip,width=\textwidth]{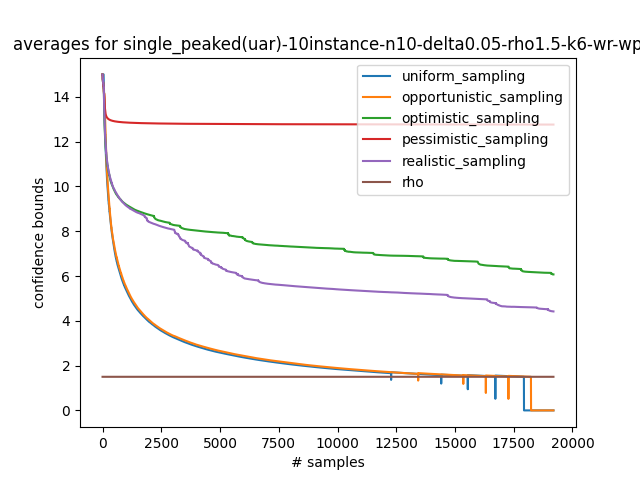}
         
     \end{subfigure}
    \caption{Sampling With Replacement, pruning applied in every step: Averages over 10 instances. $\arms=6$, $\rho = 1.5$, $n=10$, $\delta = 0.05$. Left: Instances Sampled from Mallows Model, Right: Single-Peaked Instances.}
    \label{fig:msp-wr-wp4}
 \end{figure*}

}

\section{Conclusion}
We phrased the problem of eliciting voters' preferences for determining a Kemeny ranking in terms of a Dueling Bandit problem and found that while there are no efficient PAC algorithms for approximating the Kendall tau distance, we can formulate PAC algorithms for approximating the Kemeny score. 
Here, our approximation bounds for the Kemeny score are dependent on the sum of confidence bounds for the approximated winning probabilities of arms, rendering uniform sampling the best sampling strategy w.r.t. these bounds.
We analyse the sample complexity for uniform sampling of arms for sampling with and without replacement.
Further, we described ways of pruning confidence bounds in order to reduce the sample complexity, yielding non-uniform sample strategies, and demonstrate the effect of our methods experimentally. 

In future work, one could transfer the methodology taken in this paper for finding confidence-bound-based approximation bounds for other C2 voting rules. 
Furthermore, the Kemeny score could also be an interesting approximation measure  for PAC algorithms in other settings than just Kemeny rankings.

\section*{Acknowledgements}
This work was  supported under the NRC Grant No~302203
“Algorithms and Models for Socially Beneficial Artificial Intelligence”.

\bibliography{ref}

\clearpage
\appendix
\section*{Supplementary Material}
\appendixProofText

\end{document}